\documentclass[11pt]{article}

\usepackage{iftex}
\ifPDFTeX
  \usepackage[utf8]{inputenc}
  \usepackage[T1]{fontenc}
\fi
\usepackage[margin=1in]{geometry}
\usepackage{amsmath}
\usepackage{amssymb}
\usepackage{graphicx}
\usepackage{array}
\usepackage{longtable}
\usepackage{xcolor}
\usepackage{listings}
\usepackage[numbers,sort&compress]{natbib}
\usepackage[colorlinks=true,linkcolor=blue,citecolor=blue,urlcolor=blue]{hyperref}

\graphicspath{{figs/}}

\newcommand{\figcap}[2]{%
  \par\vspace{5pt}%
  {\footnotesize\setlength{\parindent}{0pt}\textbf{#1~$\vert$~}#2\par}}

\title{Verifiable abstention makes AI leak diagnosis accountable\\
in urban water distribution networks}

\author{Tianwei Mu\textsuperscript{1,2,3},\quad Yue Wang\textsuperscript{1},\quad Mingzhe Yuan\textsuperscript{2,4,*},\quad Wenhong Wang\textsuperscript{2},\quad Qing Luo\textsuperscript{3},\quad Min Xiao\textsuperscript{3,*}\\[3pt]
Xuerui Yin\textsuperscript{1},\quad Hui Yang\textsuperscript{1},\quad Jun Li\textsuperscript{1},\quad Dan Xue\textsuperscript{6},\quad Manhong Huang\textsuperscript{5,*}}

\date{\today}

\begin{document}

\maketitle

\begingroup
\renewcommand{\thefootnote}{}%
\footnotetext{\raggedright
  \textsuperscript{1}School of Municipal Engineering and Environment,
  Shenyang Jianzhu University, Shenyang 110168, China.
  \textsuperscript{2}Guangzhou Institute of Industrial Intelligence,
  Guangzhou 511458, China.
  \textsuperscript{3}Key Laboratory of Ecological Restoration of Regional
  Contaminated Environment, Ministry of Education, College of Environment,
  Shenyang University, Shenyang 110044, China.
  \textsuperscript{4}Shenyang Institute of Automation, Chinese Academy of
  Sciences, Shenyang 110169, China.
  \textsuperscript{5}College of Environmental Science and Engineering, State
  Environmental Protection Engineering Center for Pollution Treatment and
  Control in Textile Industry, Donghua University, Shanghai 201620, China.
  \textsuperscript{6}School of Information Science and Engineering, Shenyang
  University of Technology, Shenyang 110023, China.
  \textsuperscript{*}Corresponding authors: Mingzhe Yuan,
  \href{mailto:mzyuan@sia.cn}{mzyuan@sia.cn}; Min Xiao,
  \href{mailto:xyz012263@163.com}{xyz012263@163.com}; Manhong Huang,
  \href{mailto:huangmanhong@dhu.edu.cn}{huangmanhong@dhu.edu.cn}.
  First author: Tianwei Mu,
  \href{mailto:1189233@mail.dhu.edu.cn}{1189233@mail.dhu.edu.cn}.}%
\addtocounter{footnote}{-1}%
\endgroup

\begin{abstract}
Leak localization is usually evaluated as forced-choice prediction, although sparse hydraulic observations may not justify excavation. Here, we quantify a pressure-information limit and use it to recast localization as selective, evidence-gated decision-making. A physics-grounded executor falsifies competing leak, demand, sensor and valve hypotheses in a hydraulic twin. Deterministic code computes every number and every acceptance predicate; an independent large language model auditor may add a rejection but never overturn a failed check. Forced retrieval placed only 95 of 300 leaks in the correct zone. Across 550 mixed events, the gate acted on 223 (214 correct); on a third-party 33-leak benchmark, all four accepted events were correct. In a replay of 194 audited City D repairs, the pressure tier authorized five excavation recommendations, three matching the repaired district, while the district-inflow tier returned the correct district for 85 events. Observability limits with machine-checkable abstention enable auditable utility intervention.
\end{abstract}

\section*{Introduction}

Physical leakage from pressurized mains wastes treated water and the energy used to produce it; the volume of water lost each year is estimated at 126 billion cubic meters worldwide \cite{ref1,ref2}. Locating the leak in urban water distribution networks (WDNs) remains a standing operational problem. The difficulty is structural: a leak reaches the network only as a small pressure or flow perturbation, utilities meter a few tens of points among hundreds to thousands of junctions, and the resulting inversion is ill-posed, with candidate locations far outnumbering sensors and hydraulically similar candidates indistinguishable at instrument noise \cite{ref3}. Learned localizers attack this inversion directly, from graph neural networks \cite{ref4} and burst-detecting sequence models \cite{ref5,ref6,ref7} to cross-network transfer \cite{ref8} and physics-informed surrogates \cite{ref9,ref10,ref11}.

These systems are, however, evaluated as prediction: name a location for every event and score how often it is right. The intervention they support obeys a different logic. An excavation commits a crew and cannot be undone, and the deviation that triggered it need not be a leak at all, because demand surges, sensor drift and valve mis-operation all produce leak-like signatures \cite{ref12}. The operational question is therefore not where the most likely leak lies but whether the evidence in hand justifies acting on it. Selective classification supplies a reject option \cite{ref13,ref14}, learning-to-defer a hand-off to a human expert \cite{ref15}, and conformal prediction distribution-free guarantees on coverage or on a chosen risk \cite{ref16,ref17}. Yet each abstains on a scalar score: none establishes which competing hydraulic causes have been excluded, or what evidence a particular excavation requires, and that justification is what an operator or regulator must weigh (Fig. 1).

\begin{figure}[htbp]
\centering
\includegraphics[width=\textwidth,height=0.75\textheight,keepaspectratio]{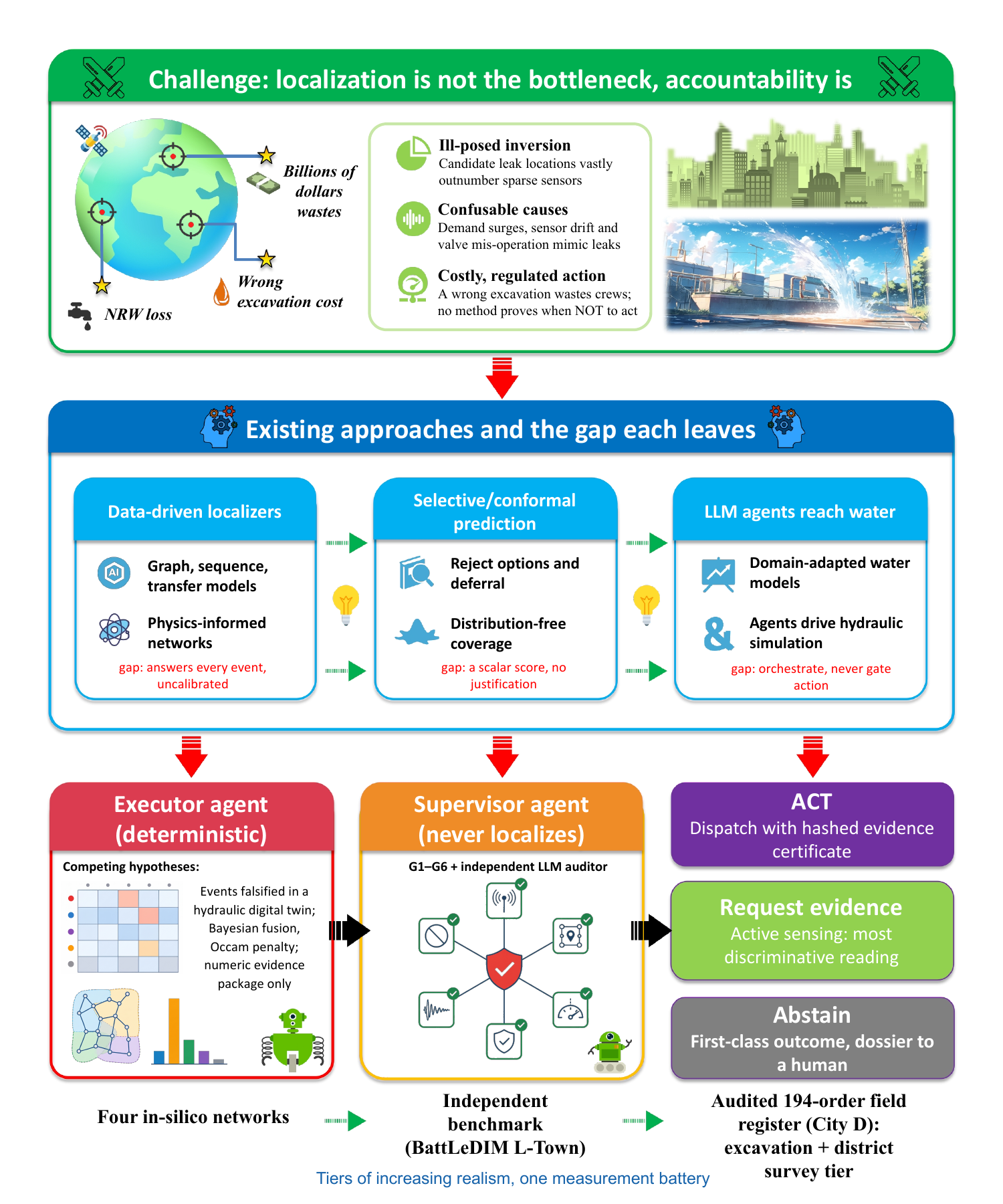}
\figcap{Fig. 1}{\textbf{The accountability gap in leak localization and the position of this work.} Top, the challenge: the inverse problem is ill-posed (candidate locations vastly outnumber sparse sensors), leak-like deviations arise from non-leak causes (demand surges, sensor drift, valve mis-operation), and a wrong excavation is costly, so a localizer that must answer every event cannot be trusted to dispatch a crew. Middle, existing approaches and the gap each leaves: data-driven localizers force an answer on every event; selective and conformal methods abstain on a scalar score without auditable justification; emerging LLM agents orchestrate simulation but do not gate action. Bottom, this work: leak diagnosis recast as accountable decision-making under verifiable abstention, in which a deterministic executor agent falsifies competing hypotheses against a digital twin and an independent supervisor agent, combining a code-verifiable goal contract of six hard predicates with a LLM auditor, certifies a dispatch with a hashed evidence certificate, requests the most discriminative additional evidence, or abstains with a dossier for a human. The bottom row names the tiers of increasing realism to which the identical measurement battery is applied: four in-silico networks, the independently generated BattLeDIM L-Town benchmark, and the audited 194-order City D repair register with its excavation and district-survey tiers.}
\end{figure}

Large language models (LLM) enter the framework introduced here only as constrained components. Agents that interleave reasoning with tool calls \cite{ref18,ref19}, their planning, memory and tool use now systematized \cite{ref20}, have begun to reach environmental and water applications as domain-adapted models and simulation-driving agents \cite{ref21,ref22,ref23,ref24}. Here they appear at two points only: an optional planner that chooses which non-leak hypothesis families receive extra scrutiny, leak zones being tested exhaustively regardless, and an independent auditor that reviews a decision its own model did not produce \cite{ref25,ref26,ref27}. Every number and every decision predicate is computed in code by the hydraulic tools, so the physics and the decision gate remain deterministic and the language models cannot loosen them.

The starting point of this study is an observability bound, not a localization algorithm. Using 194 audited repair locations from City D and the utility-provided network model, we evaluated the pressure signatures that these events would produce under a prespecified measurement-noise model. At the most informative junction, the median clean signature was 0.0023 m; only five events exceeded a 0.15 m single-measurement noise scale and one exceeded the 0.45 m (3$\sigma$) actionability threshold. Even under an idealized gauge-at-every-junction matched filter with five-night averaging, only 45 and 17 events, respectively, crossed these thresholds. For this network and severity distribution, excavation-scale dispatch is therefore constrained primarily by the information available from pressure, rather than by the choice of localization algorithm.

Here we build this limit into the decision process itself, making three contributions. We first quantify the pressure-information limit and show that it survives the strongest instrument one could posit, a gauge at every junction combined by the optimal filter, so the boundary on excavation-scale dispatch is set by the information rather than by the algorithm. We then recast leak diagnosis as accountable decision-making under verifiable abstention: a physics-grounded executor falsifies competing hypotheses in a hydraulic twin, and an independent supervisor audits the resulting evidence against six code-verifiable predicates before certifying a dispatch with a tamper-evident digest, requesting more evidence, or abstaining. Finally, the same gate delivers a graded operational response, excavation, district survey or abstention, across three tiers of realism. On EXA7 the system acts on 223 of 550 mixed leak-and-confounder events (40.5\% coverage) and names the correct zone in 214 (96.0\%, 95\% CI 92.5-98.1\%), whereas a forced retrieval localizer scores 95 of 300 leaks (31.7\%) on the same unit: precision conditional on acting, not accuracy on a shared denominator. Under the transfer gate introduced with the L-Town benchmark and carried over unchanged, with no threshold tuned on the register, it localizes all 4 of the 33 L-Town leaks it accepts (95\% CI 39.8-100.0\%), issues 5 excavation recommendations across the 194 orders (2.6\% coverage) of which 3 are correct (60\%, 95\% CI 14.7-94.7\%), and places all 85 survey-tier recommendations (43.8\% coverage) in the correct hydraulic zone (95\% CI 95.8-100.0\%). Every action and every abstention carry the record needed to review it.

\section*{Results}

\subsection*{The executor-supervisor architecture}

The executor (Fig. 2) scores a structured hypothesis space: a leak in one of K zones against a zone-wide demand anomaly, a sensor fault, a valve mis-state or no actionable anomaly. Each hypothesis is scored by whether it reproduces the observed deviations in the twin, and the survivors are fused into a Bayesian posterior with an Occam penalty against over-flexible leak fits (Eqs. 1-3). The supervisor then evaluates the goal contract (Eq. 4) whose deterministic predicates include the margin over the runner-up zone, exclusion of rival explanations, and a twin residual that physically reproduces the observation. Acceptance emits a hashed evidence certificate; rejection requests more evidence; otherwise, the system abstains, a first-class outcome rather than a forced guess.

Diagnosis runs on a fixed spatial scaffold computed offline for every network (Fig. S1), drawn in Fig. 3 for EXA7 and for City D, whose 15 districts carry 23 sensors. The zone is the operational unit, what a repair crew searches and what every number below is scored at. One identical battery of five measurements (Table 1) is applied unchanged to four in-silico networks (EXA7, KY4, the City H municipal model and the City D district model) and to two external legs, the BattLeDIM L-Town benchmark and City D's audited 2025 repair register. All six differ in information content: the system adapts its coverage to that content while defending its precision.

\begin{figure}[htbp]
\centering
\includegraphics[width=\textwidth,height=0.75\textheight,keepaspectratio]{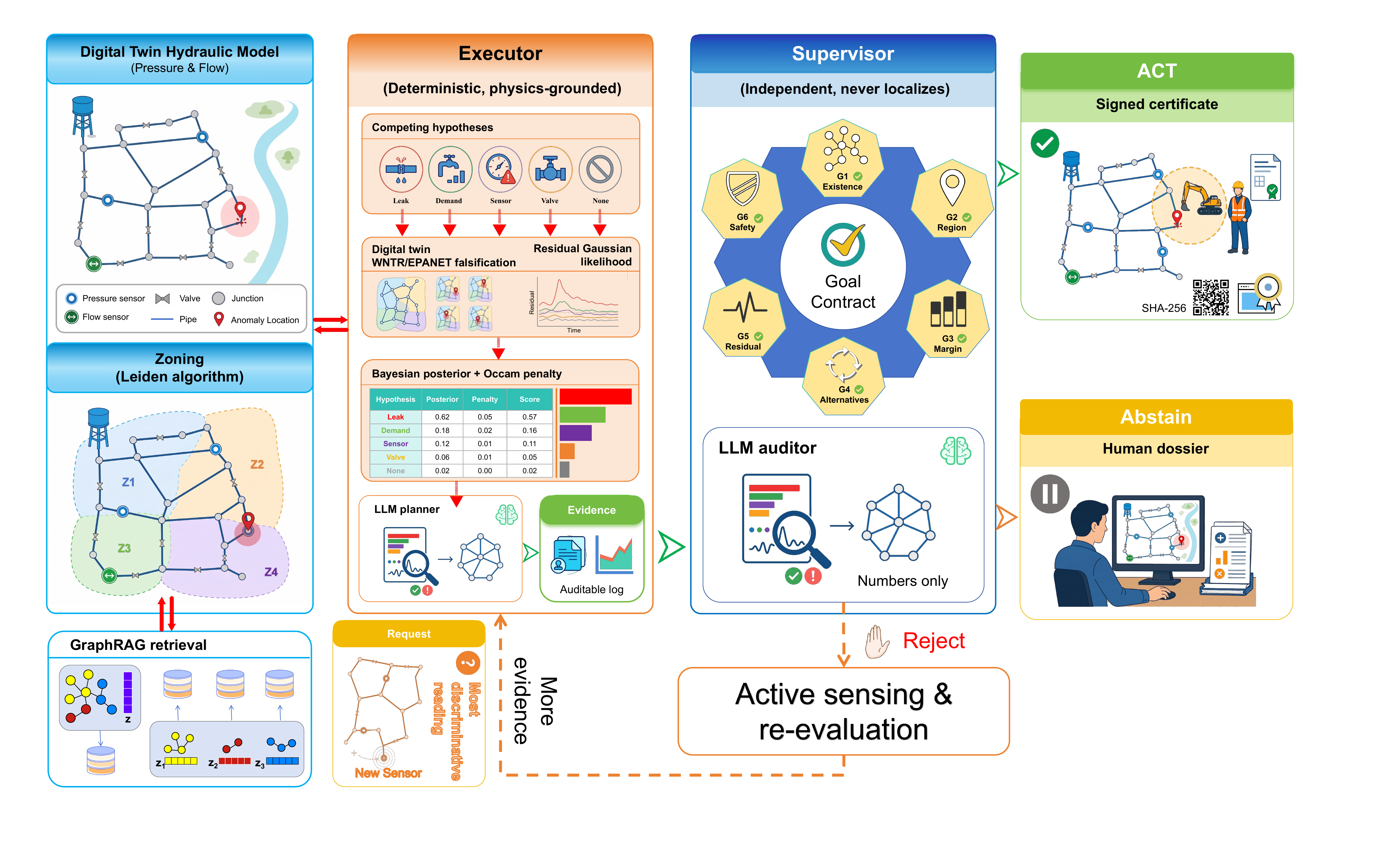}
\figcap{Fig. 2}{\textbf{Architecture of the executor-supervisor agent for accountable leak diagnosis.} Left, the offline inputs computed once per network: the WNTR/EPANET digital twin, the Leiden hydraulic zoning with its sensor placement, and the topology-aware retrieval library. Center left, the executor (orange) enumerates a mutually exhaustive set of competing hypotheses (leak, demand anomaly, sensor fault, valve mis-state, or none), realizes each in the twin, falsifies those that cannot reproduce the observation, and fuses the survivors into a single posterior; retrieval is one evidence source among several, not the answer. Center right, the supervisor (blue) never localizes: it audits the numeric evidence package against the goal contract, whose six predicates are named around the wheel and carry a tick when they pass (existence, region, margin, alternatives, residual, safety). Language enters at exactly two points, in models of different families: a planner in the executor, used in the optional dual-model configuration, which only chooses which non-leak hypotheses receive extra scrutiny and so can never cause a leak to be missed; and an auditor in the supervisor, which sees the numeric summary alone and may add a rejection but never overturn a failed hard check. Every number entering a decision comes from the physics tools. Right, the three outcomes: ACT with a certificate bound to the evidence by a SHA-256 digest, ABSTAIN with a dossier for a human, and, for a resolvable failure, a request for evidence that triggers active sensing and re-evaluation. The falsification, fusion and contract arithmetic drawn in the center panels are defined in Methods. Panel values illustrate the format of each step; all reported metrics come from the deterministic pipeline.}
\end{figure}

\begin{figure}[htbp]
\centering
\includegraphics[width=\textwidth,height=0.75\textheight,keepaspectratio]{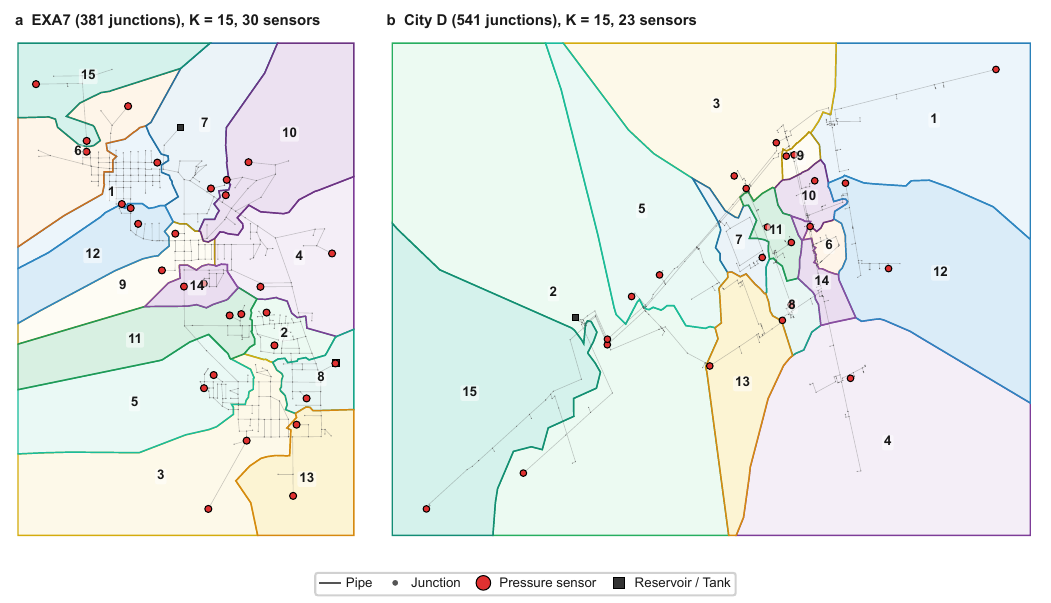}
\figcap{Fig. 3}{\textbf{Hydraulic zoning and sensor placement of the benchmark and utility networks.} Leiden partitions of the pressure-weighted modularity, drawn as space-filling territories whose colors encode zone identity only; the territory construction and the sensor-siting protocol are in Methods. a, EXA7: K = 15 zones with its 30 pressure sensors. b, City D: K = 15 districts with its 23 pressure sensors. In both panels the in-figure key names the marks: large red circles are pressure sensors, dark squares are reservoirs and tanks, and small grey dots are junctions; grey lines are pipes, and the number printed inside each territory labels its zone.}
\end{figure}

\subsection*{Selective decision quality on the EXA7 benchmark}

On EXA7 (381 junctions, K = 15 zones, 30 pressure sensors; Fig. S1a) we run the mixed test set at sensor noise $\sigma$ = 0.05 m (Fig. 4a). Forced to answer on every event, the topology-aware retrieval localizer attains only 31.7 $\pm$ 2.0\% top-1 zone accuracy (five seeds), the brittle base predictor that motivates abstention.

The executor raises zone accuracy to 81.7 $\pm$ 5.0\% (Note 7). The supervisor converts this into a decision precision of 96.1 $\pm$ 3.3\% at its operating point, acting on 40.5\% of events pooled over seeds; tightening the residual gate reaches 100\% precision at 13 $\pm$ 15\% coverage. A paired McNemar test against the forced retrieval localizer is decisive (p < 0.001; Note 7), and the contract dominates a plain existence-threshold baseline across the whole risk-coverage curve (Fig. 4a).

That dominance has a mechanism: the scalar existence score saturates, so even its strictest setting cannot buy precision, whereas the margin, alternative-exclusion and residual predicates hold the full contract at 100\% precision out to 39\% coverage (representative seed, Fig. 4a; Note 8). Where the executor and the forced retrieval localizer disagree on pooled leak events, the executor is right in 254 of 286 cases (89\%).

\begin{figure}[htbp]
\centering
\includegraphics[width=\textwidth,height=0.75\textheight,keepaspectratio]{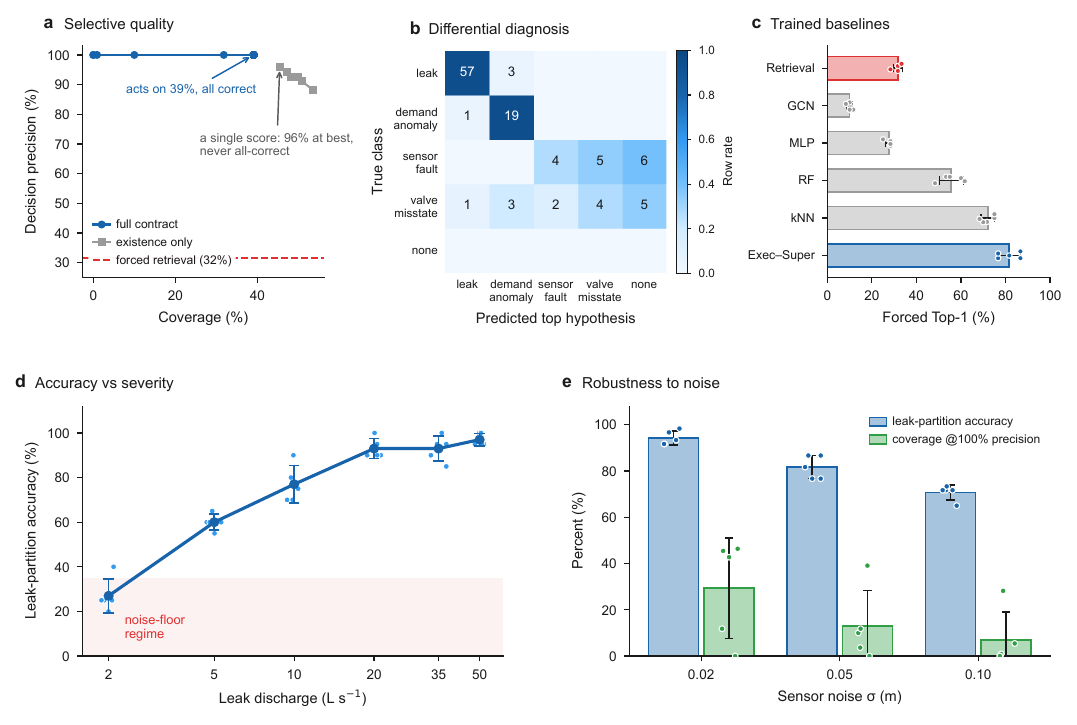}
\figcap{Fig. 4}{\textbf{Selective decision quality, differential diagnosis and forced baselines on the EXA7 benchmark. a, Decision precision versus coverage at $\sigma$ = 0.05 m for a representative seed (n = 110 events) as the acceptance threshold is swept: the multi-predicate goal contract (blue) dominates a single existence-threshold baseline (grey); the red dashed line is the mean forced-retrieval top-1 accuracy over the five seeds at coverage = 1. b, Confusion matrix of the executor's top hypothesis (columns) against the true event class (rows) for the representative seed (seed 42; n = 110 events).} Cells are event counts and the color bar is the row rate, each row normalized by its own class total; the 'none' row is empty because this mixed set contains no true no-event cases, which are scored separately as the no-leak controls of Table 1. Sensor faults are never mis-typed as leaks and demand anomalies are typed correctly in 19 of 20 cases; the corresponding five-seed pooled matrix, computed on the larger severity-sweep event set (600 leak and 250 confounder events), is in Table S1. c, Forced top-1 zone accuracy on identical held-out test leaks (n = 60 leaks per seed, 300 pooled over the five seeds): the topology-aware retrieval baseline (red), four trained localizers (grey) and the executor (blue). The executor bar is zone accuracy after supervisor-directed active sensing, 81.7 $\pm$ 5.0\%, so its error bar is that quantity's seed spread; twin-fit re-ranking alone gives 81.7 $\pm$ 2.6\% (Results). A two-sided, continuity-corrected paired McNemar test on the separate severity-sweep leak set (2-50 L s[-1], n = 600 over the same five seeds) compares the contract-gated decision, not the twin-fit ranking, against the forced retrieval localizer: $\chi$\cite{ref2} = 171, 254 versus 32 discordant events, P < 0.001. d, Leak-partition accuracy versus injected discharge on the severity-sweep set (n = 100 leaks at each of the six discharge levels, pooled over the five seeds), rising from the noise-floor regime at 2 L s[-1] to 97\% at 50 L s[-1], non-decreasing at every step. e, Leak-partition accuracy (blue bars) and the coverage attainable at 100\% decision precision (green bars) at three sensor-noise levels. Accuracy is 94.3 $\pm$ 3.0\%, 81.7 $\pm$ 5.0\% and 70.7 $\pm$ 3.2\% and coverage is 29.3 $\pm$ 21.8\%, 12.9 $\pm$ 15.4\% and 6.9 $\pm$ 12.1\% at $\sigma$ = 0.02, 0.05 and 0.10 m respectively. In c-e, bars and points show the mean over n = 5 independent random seeds, overlaid circles are the individual seeds, and error bars are the standard deviation over those five per-seed values (DDOF = 1); numerical values for c and d are in Tables S3 and S1.}
\end{figure}

\subsection*{Differential diagnosis of confounders}

The executor tests non-leak hypotheses, performing genuine differential diagnosis rather than assuming a leak (Fig. 4b). Pooled over five seeds (n = 600 leak and 250 confounder events; Table S1), demand anomalies are correctly typed in 99 of 100 cases, sensor faults are never mis-classified as leaks, and 547 of 600 leaks are typed as leaks. Typing is weakest for the discrete valve and sensor confounders, but the mis-typings land almost entirely in other non-leak classes that never dispatch a crew (Note 9). The quantity that matters, a confounder dispatched as a leak, is 1 in 250 with the differential head and 50 in 250 without it.

\subsection*{Trained baselines, calibration, and component ablations}

Trained localizers on the identical test leaks confirm the retrieval anchor is not a strawman (Fig. 4c and Table S2). The strongest, a cosine k-nearest-neighbor classifier, reaches 72.0 $\pm$ 3.0\% forced top-1; the executor's twin-fit attains 81.7 $\pm$ 5.0\% and, uniquely, converts it into a selective decision, traced to twin falsification rather than pattern coverage (Note 10). Accuracy is strongly severity-dependent, rising monotonically from 27\% at 2 L s[-1] to 97\% at 50 L s[-1] (Fig. 4d and Table S3), so abstained events are information-limited rather than arbitrary failures.

The residual-derived confidence is reasonably calibrated and conformally risk-controlled (Table S4 and Note 11). Among settings that act on at least 30\% of events, the goal contract reaches 97.9\% decision precision (140/143 acted at 32.5\% coverage), about 1.5 points below a logistic-calibrated threshold and clearly above an existence-only threshold. Unlike either, it needs no correctness labels and names the predicate behind every rejection (Note 11). Ablations confirm the predicates do real work: loosening the contract to existence-and-residual drops precision from 93.6\% to 81.2\%, and disabling the Occam penalty multiplies demand-to-leak mis-typing nine-fold (Table S5 and Note 11). A fully dual LLM configuration reproduces the deterministic accept/abstain decisions exactly (100\% agreement), adding interpretability without altering the physics-grounded decision (Note 11).

\subsection*{Abstention transfers to an independent benchmark}

To test transfer to data we did not generate we evaluated the public BattLeDIM L-Town dataset: 782 nodes, 33 pressure sensors, 33 ground-truth pipe leaks over two years (Fig. S1b). Diagnosis runs against the nominal network model, whose parameters differ from the benchmark's own data-generating model by up to 10\%, a genuine sim-to-real test.

Raw localization is hard in this regime: forced top-1 zone accuracy is only 15\% (5 of 33). The accountable system responds exactly as intended, abstaining on 88\% of events and acting on the 4 leaks it can support at 100\% decision precision with zero false dispatches (4/4; Table S6). Abstention therefore transfers to an independently generated benchmark; coverage under severe model error does not (Note 12).

Sweeping the acceptance threshold traces the full risk-coverage frontier (Fig. 5a; Table S7). At matched four-event coverage the contract reaches 100\% precision versus 75\% for a scalar existence threshold (4/4 versus 3/4; Fig. 5b): on third-party data without correctness labels, margin and alternative exclusion strictly improve precision (Note 13).

\begin{figure}[htbp]
\centering
\includegraphics[width=\textwidth,height=0.75\textheight,keepaspectratio]{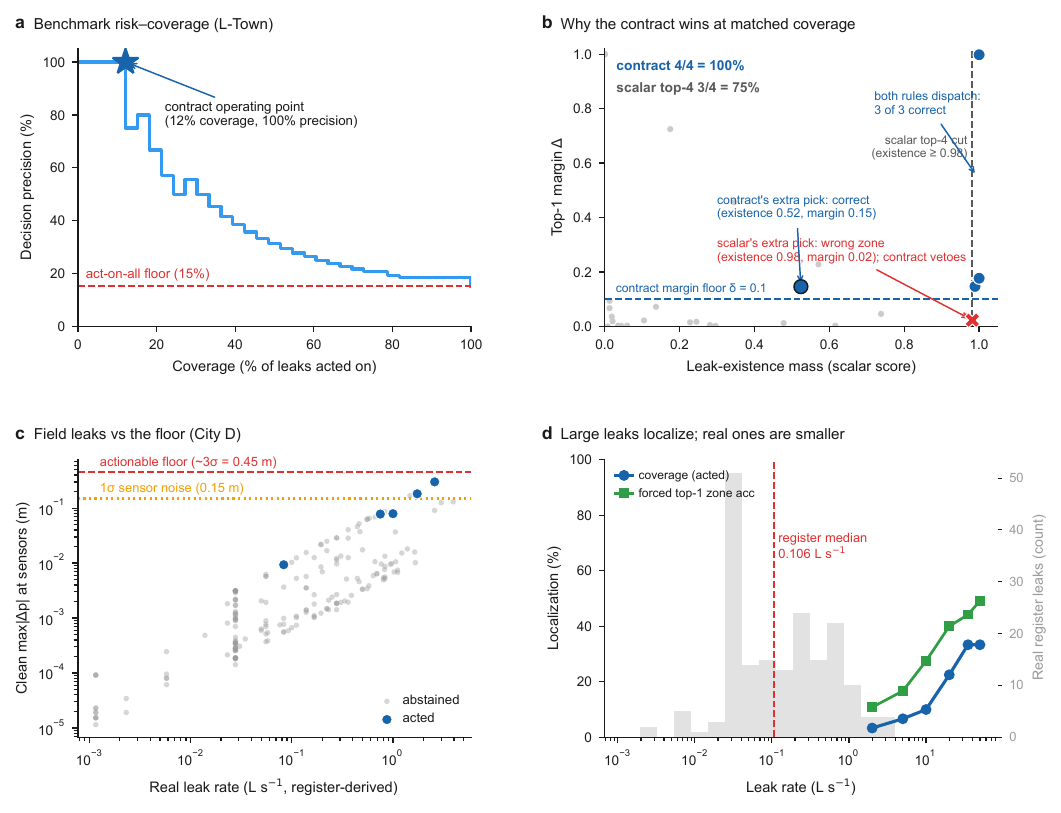}
\figcap{Fig. 5}{\textbf{The two external tiers: the BattLeDIM L-Town benchmark and the City D repair register. a, Risk-coverage on L-Town (n = 33 ground-truth leaks, competition SCADA diagnosed against the nominal model): the existence-threshold frontier runs from 100\% precision at 9\% coverage down to the 15\% act-on-all floor (red dashed); the goal-contract operating point (star; 12\% coverage, 100\% precision) lies above the frontier at matched coverage. b, The mechanism at matched four-event coverage, shown in the existence-margin plane.} Small grey dots are the n = 33 L-Town leaks, large blue circles are the four events the goal contract dispatches, and the red cross is the fourth event picked by the scalar rule and vetoed by the contract. The scalar rule ranks by leak-existence mass alone (grey dashed vertical cut) and its fourth pick, a high-existence event with margin 0.02, goes to the wrong zone; the contract's margin floor ($\delta$ = 0.10, blue dashed horizontal line) vetoes that event and admits a cleanly separated lower-existence event instead (existence 0.52, margin 0.15). Three events are dispatched by both rules and all three are correct, so the contract reaches 4/4 (100\%) against the scalar rule's 3/4 (75\%). c, Each of n = 194 audited 2025 City D work orders, plotted as register-derived leak rate against its clean peak sensor deviation (log-log): every event lies below the $\approx$ 3$\sigma$ actionability floor (0.45 m, red dashed; maximum 0.30 m) and all but three below the 1$\sigma$ noise level (0.15 m, amber dotted); the transfer-preset contract dispatches five events (blue; three correctly) and abstains on the rest (grey), while the stricter audit preset abstains on all 194; 50 no-leak controls yield one false dispatch under the transfer preset and none under the audit preset (Methods). d, Controlled severity sweep on the same network (single leaks at 120 seeded nodes, 2-50 L s[-1]): coverage (circles) and forced top-1 accuracy (squares) versus leak rate, with the register's severity distribution overlaid (grey histogram, right axis; median, red dashed); the median signature clears the noise floor only near 10-20 L s[-1], far to the right of the entire register. City D node assignments are audited real locations; pressures are twin-simulated. Per-leak and per-band values are in Tables S6, S7, S11 and S12.}
\end{figure}

\subsection*{An audited repair register on the utility's own network: City D}

The operating utility of City D, a district network in China, provided its own EPANET model (541 junctions, 79 throttle-control valves, used unmodified; Fig. S1d) and its 2025 repair register of 256 leak work orders, of which a strict inclusion filter admits 194 at audited real locations. Each signature is twin-simulated at its audited node with field-scale noise ($\sigma$ = 0.15 m).

The register's severity distribution is unforgiving: the 194 orders have a median loss rate of 0.106 L s[-1] and a median clean signature of 0.002 m, two orders of magnitude below the 0.15 m noise floor, and no event reaches the 0.45 m actionability floor (Fig. 5c and Table S8). The accountable system dispatches excavation on five events (2.6\% coverage), three of them correctly (60\% acted precision), and raises one knife-edge false dispatch on 50 no-leak controls, dismissed by the stricter audit preset, the same contract at higher existence and margin floors (Note 14). A forced localizer would have dug 194 times at 12\% precision; the contract digs five times at 60\%.

A controlled severity sweep confirms this reflects the leak sizes and hydraulics rather than a pipeline failure (Fig. 5d and Table S9): injected leaks clear the noise floor only near 10-20 L s[-1] while the entire register lies below 4 L s[-1], and under the standard in-silico protocol the same network supports 81.8\% pooled precision at 20.0\% coverage (Table 1 and Note 15). The register's small-leak majority therefore needs a complementary modality.

Pressure evidence is bounded here. With a sensor at all 541 junctions, the single best gauge puts only 5 of 194 events above the 1$\sigma$ noise level; even the matched filter across the whole array, granted the survey tier's own 5-night averaging, reaches 3$\sigma$ on just 17 (Table S10). The contract therefore adds a coarser second tier on a different physical quantity, district inflow, which by mass balance rises by the leak rate itself. A survey dispatch to the district with the largest observed nightly rise, gated at 3.4$\sigma$ after correction for City D's 15 districts, recovers 85 of 194 register events (44\% coverage) at 100\% district precision, with zero false dispatches on 50 no-leak control campaigns (Note 16). The survey tier sends an acoustic crew to the right district, the modality this register demands.

\subsection*{Robustness, generalization and auditable autonomy}

The system degrades gracefully as noise rises (Fig. 4e): zone (leak-partition) accuracy falls from 94.3 $\pm$ 3.0\% ($\sigma$ = 0.02 m) through 81.7 $\pm$ 5.0\% (0.05 m) to 70.7 $\pm$ 3.2\% (0.10 m). Decision precision, by contrast, moves only from 98.1 $\pm$ 1.9\% through 96.1 $\pm$ 3.3\% to 91.3 $\pm$ 3.4\%: the gate pays almost entirely in coverage. The headline metrics are stable across seeds (Fig. S2), and the gate is not delicately tuned: pooled precision stays within 89.9-99.1\% across 28 threshold settings (Table S11 and Note 17). The behavior transfers under the identical protocol to the public Kentucky-database network KY4 \cite{ref28}, where the supervisor acts on 34.4 $\pm$ 1.7\% of events at 96.4 $\pm$ 2.9\% precision, and to City H and City D (Fig. S1c,e; Table 1, Table S12): precision thins with information content, yet every gate acts far above its forced base rate.

When the supervisor rejects an event for unresolved zone ambiguity, it requests the most discriminative additional measurements (three hidden nodes per round) and re-ranks. Mean zone accuracy on the 5-50 L s[-1] headline leak set is unchanged (81.67\% before and after), and the benefit appears on the extended 2-50 L s[-1] set (70.5\% to 74.5\%; Table S5 and Note 18): active sensing is the designed recourse for the low-coverage regime, not a universal gain.

Finally, accountability is tested rather than asserted: an independent LLM supervisor (deepseek-v4-pro, a different family from any executor component) audits each decision's numeric summary alone. In a corruption stress test, it caught 16/16 deliberately corrupted evidence packages and raised no false alarm on 16 genuine ones (0/16), uniformly across corruption classes and stable across repeats (Table S13 and Note 19). It is also more than a rule-checker. Four further corruption classes pass every enumerated prompt rule individually yet cannot jointly hold. On these, a literal five-rule checker run as a control matches every auditor on the 48 in-taxonomy corruptions and raises no false alarm, yet by construction rejects 0 of 64. The model given the same prompt catches 0 to 2; one added clause, inviting any other internal inconsistency, lets it reject 49 by explicit judgement (Note 19). This is the one place the LLM is load-bearing rather than ornamental, and only when invited: elsewhere it is optional, the dual-model executor reproducing the deterministic decisions exactly. It may still only tighten the hard predicates, never replace them. The auditing layer is non-collusive by construction, the prerequisite for delegating action on regulated infrastructure.

\section*{Discussion}

This study reframes leak diagnosis from naming a location to justifying an intervention: taking the weakness of simulation-grounded localization under field-scale noise as a premise, we ask when to act and how to prove it, not how to raise a forced-choice score. The executor-supervisor agent unites four known ingredients, new in combination: twin falsification, differential diagnosis of confounders, a code-verifiable abstention contract, and an independent auditor. On EXA7, where labels can calibrate a threshold, the contract is not superior, trailing that threshold at matched coverage (97.9\% versus 99.4\%); it improves decisions only where labels are absent (Note 20). This sits within selective prediction and learning-to-defer \cite{ref13,ref15} but supplies justification that scalar rules lack; the closest agentic water system never gates action  \cite{ref23,ref24}.

End-to-end provenance checks enforce consistency among hydraulic tool outputs, posterior updates, evidence packages and reported decision predicates. For the version reported here, all in-silico experiments were regenerated through the committed pipeline, and every figure and table was produced from the resulting machine-readable records. This provenance discipline illustrates why auditability must be enforced throughout a safety-critical architecture rather than added only after diagnosis \cite{ref29}.

The architecture is domain-agnostic over networked physical-fault diagnosis wherever an offline-simulatable twin, spatially structured signatures, confusable confounders and costly actuation coincide, as in power-system faults \cite{ref30}, gas pipelines \cite{ref31}, contamination-source identification \cite{ref32}, structural-health monitoring \cite{ref33} and data-scarce water-quality prediction \cite{ref34,ref35}. We demonstrate it on water networks and offer the generalization as a hypothesis.

Several boundaries remain. The confounder classes are drawn from literature priors. The strongest external test is L-Town, whose SCADA the competition organizers generated with an independent, deliberately perturbed copy of the network model, and where the system acts at 100\% precision but on only four events. A published state-of-the-art localizer is not run head-to-head: BattLeDIM entrants natively solve a detection-and-localization protocol that does not define the selective dispatch quantities compared here, so the trained localizers of Fig. 4c serve as the forced comparison; conformal exchangeability can break under sim-to-real shift; active sensing's benefit at this sensor density is small; the LLM executor contributes interpretability rather than accuracy; and the LLM auditor generalizes beyond its enumerated rules only when invited to and only partly (Table S14), so it complements the hard predicates rather than substituting for them. A deployment on live SCADA, where the certificate trail meets a real control room and the twin-based evidence model meets measured telemetry for the first time, is the decisive next step.

\section*{Methods}

\subsection*{Problem formulation: from localization to accountable differential diagnosis}

We represent a water distribution network as a graph $\mathcal{G}=(\mathcal{V},\mathcal{E})$ with $N=|\mathcal{V}|$ junction nodes and $M=|\mathcal{E}|$ pipe links. Under steady-state operation the hydraulic state obeys conservation of mass at every node and the Hazen--Williams head-loss relationship along every pipe (Note 1, Eqs. S1 and S2). Any anomaly, whether a leak, a demand surge, a sensor fault or a mis-operated valve, induces a sensor-space pressure deviation relative to the no-anomaly baseline,

\begin{equation}
\Delta\mathbf{p}^{\mathrm{obs}} = \Delta\mathbf{p}_{\mathcal{S}}^{(h)} + \boldsymbol{\epsilon},\qquad \boldsymbol{\epsilon}\sim\mathcal{N}(\mathbf{0},\sigma^2\mathbf{I}),
\end{equation}

where $\Delta\mathbf{p}^{\mathrm{obs}}$ is the observed sensor-space pressure deviation, $\Delta\mathbf{p}_{\mathcal{S}}^{(h)}$ the deviation that anomaly hypothesis $h$ induces at the sensor set $\mathcal{S}\subset\mathcal{V}$, and $\boldsymbol{\epsilon}$ zero-mean Gaussian field noise of standard deviation $\sigma$; the observation exists only at the $|\mathcal{S}|=m\ll N$ sensor locations. We do not invert Eq. 1 for a leak location, for three concrete reasons: the inversion is ill-posed, with candidate locations vastly outnumbering the $m$ sensors; the anomaly class is itself unknown, because demand surges, sensor faults and valve mis-states produce leak-like deviations; and the implied action is a costly excavation that a forced guess of 30 to 40\% accuracy cannot justify. We therefore treat the anomaly type as unknown, and replace point prediction with an auditable, safety-checked decision that may also \textit{abstain}. The unit of evaluation is therefore the action, not the prediction. The remaining methods construct an executor that performs the diagnosis and an independent supervisor that gates the action.

\subsection*{Hydraulic zoning and sensor placement}

Two upstream constructions, reused unchanged from the benchmark configuration, reduce the ill-posed inversion of Eq. 1 to a tractable problem. First, the nodes are partitioned into $K\ll N$ hydraulically coherent zones $\{C_1,\dots,C_K\}$ by maximizing a pressure-weighted modularity in the resolution-parameterized form of Reichardt and Bornholdt \cite{ref36}, $Q(\gamma)=\tfrac{1}{2W}\sum_{i,j}\big[w_{ij}-\gamma\,s_i s_j/2W\big]\,\delta(\pi(i),\pi(j))$, which reduces to standard modularity at $\gamma=1$. The edge weight $w_{ij}=T^{-1}\sum_{t}(p_i(t)+p_j(t))/2$ averages the two endpoints' nodal pressure over the $T$ time steps of one baseline hydraulic run, and is defined only on the network's own links $(i,j)\in\mathcal{E}$, pipes together with pumps and valves; $s_i=\sum_j w_{ij}$ is the node strength and $W=\tfrac12\sum_i s_i$ the total weight. Restricting weights to existing links ensures that only hydraulically coupled nodes share a zone: two junctions with similar pressures but no connecting link receive no edge and are never merged. The weights are then rescaled linearly to $[0,1]$ with a floor of $10^{-3}$, so that networks whose pressures differ by an order of magnitude are partitioned on the same numerical footing. At each resolution the objective is optimized by the Leiden algorithm \cite{ref37}, which guarantees connected communities. Since $\gamma$ controls granularity rather than the zone count directly, we sweep 300 values over $[0.001,5]$, collect the distinct partitions obtained, and retain the one whose zone count is closest to the target, $K=15$ for EXA7 and City D and $K=25$ for the three larger networks; the retained resolutions are $\gamma=0.88$, $0.60$, $0.77$, $1.17$ and $1.41$ for EXA7, City D, KY4, City H and L-Town. A fallback that repeatedly merges the most strongly connected pair of communities is available for target counts the sweep does not reach, and was not required for any network reported here. Localizing to a zone, rather than to an exact node, matches both the granularity at which a repair crew operates and the information recoverable from sparse, noisy sensors.

Second, the $m$ sensors are placed by a partition-aware greedy rule \cite{ref38,ref39} that maximizes the minimum within-zone sensitivity, $\mathcal{S}_k^{*}=\arg\max_{\mathcal{S}_k}\min_{v\in C_k}\max_{s\in\mathcal{S}_k}\big|\partial p_s/\partial d_v\big|$, so that every zone carries a distinguishable pressure signature and none is unobservable. For EXA7 these constructions yield $K=15$ zones and $m=30$ sensors (two per zone), computed once offline and held fixed throughout diagnosis; the resulting zone maps of the benchmark and field networks are shown in Fig. 3, and those of L-Town, City H and KY4 in Fig. S3.

The scaffold is built from Leiden zones of the pressure-weighted modularity and the sensor placement inside them; all five topologies are drawn in Fig. S1, EXA7's 15 zones carrying two sensors each and City D's 15 districts the 23 detection-coverage-optimized sensors. The L-Town benchmark is third-party data throughout.

\subsection*{Competing hypotheses and digital-twin falsification}

Given the zones and sensors of the preceding subsection, diagnosis is posed over a structured, mutually exhaustive hypothesis space $\mathcal{H}$ spanning one dominant anomaly per event: a leak somewhere in any zone $C_k$, a zone-wide demand anomaly in any zone, a fault on any sensor channel, a mis-state of any monitored asset, or the null hypothesis $h_0$ that no actionable anomaly exists (Note 2, Eq. S3 and Fig. S4). Crucially, the space contains explicit \textit{non-leak} explanations, so the diagnosis can conclude ``this is not a leak'' rather than being forced to localize one. Promoting the existence question to a first-class hypothesis is what makes abstention principled rather than a confidence threshold.

Each hypothesis is scored not by similarity to stored exemplars but by whether it can \textit{reproduce the observation} when realized in a calibrated hydraulic digital twin. For a hypothesis $h$, the twin (a WNTR/EPANET solver of Eqs. S1 and S2) injects the corresponding perturbation and returns a predicted sensor response $\boldsymbol{\mu}_h$. A leak is realized as a constant added demand at a candidate node, a demand anomaly as a zone-wide multiplier, and a valve mis-state as a pipe closure, each time-averaged over the simulation horizon. The residual and its Gaussian log-likelihood under the field-noise model of Eq. 1 are

\begin{equation}
\mathbf{r}_h=\Delta\mathbf{p}^{\mathrm{obs}}-\boldsymbol{\mu}_h,\qquad
\log\mathcal{L}(h)=-\tfrac{1}{2\sigma^2}\lVert\mathbf{r}_h\rVert_2^2 - m\log\!\big(\sigma\sqrt{2\pi}\big).
\end{equation}

where $\mathbf{r}_h$ is the residual of hypothesis $h$, $\boldsymbol{\mu}_h$ its twin-predicted sensor response, $\sigma$ the sensor-noise standard deviation of Eq. 1 and $m$ the number of sensors. A hypothesis that cannot reproduce the observation accumulates a large residual and its likelihood collapses; this is the falsification mechanism that distinguishes the method from retrieval. We summarize residual quality by the per-degree-of-freedom Mahalanobis distance $\rho_h=\lVert\mathbf{r}_h\rVert_2/(\sigma\sqrt{m})$, for which $\rho_h\!\approx\!1$ indicates a hypothesis consistent with the noise floor and $\rho_h\!\gg\!3$ a physically incompatible one. The null hypothesis predicts $\boldsymbol{\mu}_{h_0}=\mathbf{0}$, so it is favored only when the observation is itself within noise, the quantitative basis for ``no actionable anomaly''. A sensor-fault hypothesis $h_{\mathrm{sen}}(s)$ is handled in measurement space: it predicts the observed value on the faulty channels and zero hydraulic deviation elsewhere. Its likelihood is therefore high only when a few channels are anomalous while their hydraulic neighbors read near zero. This is the signature that separates an instrument failure from a leak, which instead depresses many sensors at once.

For the leak class, a single zone candidate is insufficient because the true leak may sit at any junction in the zone. We therefore fit each leak hypothesis by the best node--rate pair within the zone, minimizing the residual over all junctions $v\in C_k$ and the discharge grid $\mathcal{Q}=\{2,5,10,20,35,50\}$ L\,s$^{-1}$ (Note 3, Eq. S4). Because the true leak node is then a candidate, the correct zone fits its own response near-exactly, which sharply reduces cross-zone confusion relative to a coarse retrieval ranking. All twin evaluations are memoized, and a per-zone response library is built once and reused across runs, so this search over all zones is tractable.

\subsection*{Bayesian fusion and the executor agent}

Hypotheses are combined into a posterior in log-space, accumulating the prior and the per-hypothesis likelihood while correcting for the differing flexibility of the hypothesis families,

\begin{equation}
\log P(h\mid\Delta\mathbf{p}^{\mathrm{obs}}) = \log P_0(h) + \log\mathcal{L}(h) - \tfrac{1}{2}\,k_h\log m \;-\;\log Z.
\end{equation}

where $P_0(h)$ is the prior of hypothesis $h$, $\mathcal{L}(h)$ its twin likelihood from Eq. 2, $k_h$ the number of fitted free parameters, $m$ the number of sensors and $Z$ the normalizer that makes Eq. 3 a softmax over $\mathcal{H}$. The prior distributes a base rate per family (leak $0.50$, demand $0.20$, sensor $0.15$, valve $0.10$, null $0.05$) uniformly within the family. The term $\tfrac{1}{2}k_h\log m$ is a Bayesian-information-criterion (Occam) penalty in which $k_h$ counts the fitted free parameters of a hypothesis: two for a leak (node and rate), one for a demand anomaly (the multiplier), zero for a discrete valve or null hypothesis, and the number of explained channels for a sensor fault. This penalty is essential: without it, the leak family, which searches over many node--rate combinations, can spuriously out-fit a genuine zone-wide demand anomaly at sparse sensors (quantified by ablation in Table S5). From the posterior we read the three quantities the supervisor will gate on (Note 4, Eq. S5): the total leak-existence mass $\Pi_{\mathrm{exist}}$ (the summed leak posterior), the margin $\Delta$ of the top hypothesis $h_{(1)}$ over the best alternative $h_{(2)}$ of \textit{any} family, and the posterior entropy. Every term entering Eq. 3 originates from a tool output and is recorded in an evidence log, so each change in the posterior is traceable to a specific computation. This is a structural property that makes the diagnosis auditable.

The executor agent is what orchestrates these tools into the posterior of Eq. 3. It is deterministic and physics-grounded, and it never itself produces the numbers that enter the posterior, the architectural guard against fabricated metrics. In each round it first seeds the competing hypotheses: a leak hypothesis for every zone, a zone-wide demand hypothesis for each retrieved zone, a sensor-fault hypothesis over channels whose deviation exceeds four noise standard deviations, valve hypotheses on pipes incident to the leading zone, and the null hypothesis. Leak candidate nodes are the union of pre-simulated representative nodes and the highest-degree junctions of each zone.

The executor then invokes a typed tool suite: a residual-analysis tool that evaluates Eq. 2, a digital-twin tool that computes $\boldsymbol{\mu}_h$ by forward simulation, a Bayesian-update tool that applies Eq. 3, and a retrieval tool. The retrieval tool is the topology-aware Graph-RAG localizer of the prior framework, but it is used as one evidence source rather than as the answer that proposes candidate zones. Its score is not trusted as a probability; it competes, through the twin and Eq. 3, against the non-leak hypotheses. This demotion keeps the system safe when retrieval is weak under field noise: an uncertain retrieval channel leaves $\Pi_{\mathrm{exist}}$ low, so the supervisor abstains rather than forcing the retrieval top-one.

The executor finally writes a structured evidence package: the posterior over $\mathcal{H}$, the fitted leak node and rate, each hypothesis's residual and falsification status, the candidate region, and the per-tool evidence log. This package is the sole interface to the supervisor. For every hypothesis its schema fixes the family label, the fitted parameters, the predicted-versus-observed residual with its Mahalanobis distance, the posterior probability, and an explicit falsification status with the evidence reference that established it. By construction the package carries no free-text argument and no self-reported confidence, only quantities derived from the twin and Eq. 3. This is what lets a downstream auditor re-derive the decision from the numbers alone, not from a persuasive narrative.

\subsection*{The supervisor: goal contract, abstention, and active sensing}

The supervisor never localizes. It audits the executor's evidence package against an explicit goal contract (Fig. S5), a conjunction of acceptance predicates that a dispatch must satisfy. Writing $h_{(1)}$ for the top hypothesis and $\rho_{(1)}$ for its residual Mahalanobis distance, the contract authorizes an action only if every data-dependent predicate holds,

\begin{equation}
\mathrm{ACT}\;\iff\;\mathrm{G1}\wedge\mathrm{G2}\wedge\mathrm{G3}\wedge\mathrm{G4}\wedge\mathrm{G5}\wedge\mathrm{G6}.
\end{equation}

where G1 to G6 are deterministic acceptance predicates over the evidence package, as follows. \textbf{G1 (existence)}: $h_{(1)}=h_{\mathrm{leak}}$ and $\Pi_{\mathrm{exist}}\ge\tau_{\mathrm{e}}$, the top hypothesis is a leak and the summed leak-existence mass clears its threshold, which prevents action while the null, demand or sensor explanations remain plausible. \textbf{G2 (region)}: $|R|\le R_{\max}$, the candidate region is small enough for a repair crew to search. \textbf{G3 (margin)}: $\Delta\ge\delta$, the top hypothesis leads the best alternative of any family, not merely the second-best zone, by a working margin. \textbf{G4 (alternatives)}: $P(h_{(2)})\le\alpha$, the strongest rival explanation is itself improbable. \textbf{G5 (residual)}: $\rho_{(1)}\le\rho_{\max}$, the fitted leak physically reproduces the observation, with $\rho_{\max}=3$ per degree of freedom throughout. \textbf{G6 (safety)}: $\mathcal{U}=\varnothing$, the set of unsafe implied actions is empty, blocking any recommendation that implies an out-of-bound operation. The remaining thresholds are fixed per experiment before evaluation. The EXA7 selective evaluation uses $\tau_{\mathrm{e}}=0.5$, $\delta=0.12$, $\alpha=0.20$ and $R_{\max}=60$ nodes; the residual gate is swept to trace the precision--coverage curve, whose maximum-coverage end is the reported operating point. The L-Town benchmark and the City D register leg use $\tau_{\mathrm{e}}=0.5$, $\delta=0.10$, $\alpha=0.30$ and $R_{\max}=80$, reflecting their larger zones and severe model error; the City H, City D and KY4 standard in-silico legs of Table 1 and Table S12 use the EXA7 gate unchanged ($\tau_{\mathrm{e}}=0.5$, $\delta=0.12$, $\alpha=0.20$, $R_{\max}=60$). The language-model audit experiments use $\tau_{\mathrm{e}}=0.7$, $\delta=0.15$ and $\alpha=0.20$. The numbering follows the implementation's evaluator, in which every predicate of the contract is evaluated on every event. The contract structure additionally declares two boolean deployment hooks that this study does not evaluate and that carry no predicate number, a calibration-validity flag and a human-review flag. Every leg here does use the calibration matched to its own network and noise level, but that is a property of how the runs were configured, not a condition the contract checks. These predicates are deterministic arithmetic over the evidence package and constitute the system's hard guarantees: a code-level property of the decision rule, enforced on every event, not a statistical bound on error rates. On satisfaction, the supervisor emits an acceptance certificate that records the decision, the contract version and thresholds, the measured value and pass or fail of each predicate, and the accepted hypothesis and region. A cryptographic digest binds the certificate to a canonical serialization of the evidence package, so any later tampering is detectable. This is the auditable artifact a utility or regulator requires.

When any predicate fails, the supervisor does not silently force a guess. It returns a structured request identifying the unmet predicate; for example, a low margin between two adjacent zones triggers a ``separate these zones'' request that the executor must address in a further round. If the contract cannot be met within the round budget, the system \textit{abstains}: it emits a best-effort dossier (the surviving hypotheses and the reason each predicate failed) and defers to a human. Abstention is reported as a first-class outcome, never converted to a forced prediction. This is the formal mechanism by which an intrinsically weak base localizer becomes a high-precision decision system, acting only where the evidence supports it.

When a rejection reflects an unresolved zone ambiguity rather than a true absence of evidence, the executor seeks the most informative additional measurements. Among the surviving leak hypotheses, it scores each candidate hidden node by the spread of the hypotheses' twin-predicted responses at that node (Note 5, Eq. S6), a digital-twin discriminability criterion that approximates the measurement of maximal expected information gain \cite{ref40}, and reveals the three highest-scoring nodes: the nodes at which the competing zones most disagree and therefore those whose readings are most decisive. Each revealed node's value is drawn from the full-field response plus measurement noise, the sensor set is augmented, and the surviving leak hypotheses are re-fit and re-ranked over the augmented set (Eq. S4). Because only the leak family is re-fit, active sensing may only move posterior mass within that family: the leak-existence mass set by Eq. 3 is held fixed and redistributed over the re-fitted leak hypotheses in proportion to their new within-family posteriors, the non-leak posteriors are unchanged, and the summary quantities of Eq. S5 (top hypothesis, margin over the best alternative of any family) are then re-derived from the one resulting posterior, so the evidence package the supervisor and the auditor read remains self-consistent. On the benchmark, where full per-zone candidate coverage already yields strong base localization, the average benefit of this step is modest and event-dependent; it is the designed recourse for the low-coverage, high-noise regime where base localization is weak, and we report its effect without overstatement.

The Occam penalty stops an over-flexible leak fit from masquerading as a demand anomaly, and the margin predicate is evaluated against the runner-up zone. The goal contract is the formal conjunction of Eq. 4, a conjunction of deterministic predicates on leak existence, region size, margin, alternatives, physical reproduction and safety; the language-model auditor may add a rejection but never overturns a failed hard check.

\subsection*{Independent LLM audit}

Because the executor is deterministic, the language model serves only as an \textit{independent} supervisor auditor, a configuration that maximizes rather than weakens independence. The auditor receives only the structured numeric summary: the top hypothesis and its residual, the existence mass, the margin, and the alternatives with their residuals and falsification status. It never sees the executor's internal computation, prose, or any confidence adjective. This information asymmetry, together with hard predicates (Eq. 4) evaluated in code the auditor cannot override, makes the audit non-collusive by construction.

The auditor is a frontier open model of a different family from any executor component. It returns a strict-JSON accept-or-reject verdict and may only add a rejection on top of the deterministic contract, never overturn a failed hard check. It is configured to reject any internally inconsistent package, a condition verifiable from the numbers alone; the auditor's full system prompt, the numeric evidence-summary schema it receives and the strict-JSON verdict template with its conservative parse-failure default are reproduced verbatim, together with the executor-planner's planning template, in Supplementary Methods S6.

We stress-test this layer by injecting deliberately corrupted packages and measuring the catch rate, the false-alarm rate on genuine packages, and the agreement across repeated temperature-zero audits. The taxonomy spans the three ways a diagnosis can be untrustworthy yet superficially confident. An \textit{unsupported assertion} pairs a high-probability leak with a physically incompatible residual. A \textit{fabricated exclusion} declares an alternative falsified although its residual fits the observation. \textit{Inflated evidence} presents a low existence mass or a vanishing margin as decisive. Each is detectable from the numeric summary alone, and none would be caught by a threshold on a single confidence scalar. An auditor that rejects these while accepting genuine packages, reproducibly across repeats, supplies the independent second signature that turns an automated recommendation into an accountable one. A follow-up test separates rule-following from judgement: the same sixteen genuine packages receive four further corruptions that pass every enumerated prompt rule individually but cannot jointly hold, and three auditors are compared on identical inputs, the five prompt rules implemented literally in code, the language model with the prompt verbatim, and the language model with one added clause asking it to reject any other internal inconsistency without naming any; the paper's auditor and, as a second opinion, the planner's model family are both run, and every call is committed as a transcript (Table S14). These audit experiments run at the stricter audit preset ($\tau$\_e = 0.7, $\delta$ = 0.15, $\alpha$ = 0.20; 'The supervisor'), not at the headline selective gates.

\subsection*{Scenario generation, evaluation protocol, and baselines}

To evaluate the system, we generate all four anomaly classes with one seeded WNTR/EPANET pipeline, so results are bit-reproducible. Leaks are injected at representative and randomly chosen junctions across discharge rates of 5 to 50 L\,s$^{-1}$. Demand anomalies are zone-wide multipliers matched in magnitude to the leak bands, so that they are genuinely confusable. Sensor faults apply bias, stuck, or excess-noise corruptions to one to three channels of an otherwise normal state. Valve mis-states close pipes, and closures that disconnect the network are detected and discarded. The same field-scale noise ($\sigma=0.05$ m unless swept) is applied to every class, and leak test nodes are drawn disjoint from the retrieval library to prevent memorization. The identical measurement battery of Table 1 (forced top-1, coverage, acted precision, 50 no-leak controls and the matched-coverage contract-versus-scalar comparison) is applied to every network; the in-silico control and matched-coverage cells are produced by one shared evaluation routine, released with the code, under the same per-seed operating gates as the headline runs.

The unit of analysis is the action. We report risk--coverage and precision--coverage curves \cite{ref13}, together with the false-dispatch rate, the four-class confusion matrix, and the expected calibration error of the residual-derived confidence. Writing $\mathcal{A}$ for the acted events and $\hat{z}_a,z_a$ for the predicted and true zones of event $a$, decision precision is $|\{a\in\mathcal{A}:h_{(1),a}=h_{\mathrm{leak}}\wedge\hat{z}_a=z_a\}|/|\mathcal{A}|$ and coverage is $|\mathcal{A}|/n$. The false-dispatch rate is the complement of decision precision, so a non-leak event acted upon, or a leak sent to the wrong zone, both count as failures. Sweeping the residual gate $\rho_{\max}$ traces the precision--coverage curve, and the calibration error aggregates, over confidence bins, the absolute gap between empirical accuracy and mean confidence. Headline quantities are reported as mean and standard deviation over five random seeds, with a noise sweep over $\sigma\in\{0.02,0.05,0.10\}$ m. Significance against the forced localizer is assessed by a two-sided, continuity-corrected paired McNemar test on the leak subset pooled over the five seeds of the severity-sweep set (2--50 L\,s$^{-1}$, n = 600); small-sample proportions are reported with exact 95\% Clopper--Pearson confidence intervals.

As forced (coverage-one) comparators we train genuine localizers on the scenario library and evaluate them on identical test leaks: a cosine $k$-nearest-neighbor classifier, a random forest, a multilayer perceptron, and a graph convolutional network. As selective-prediction comparators we add a logistic-calibrated confidence threshold, fit on a held-out development seed, and a split-conformal selective predictor \cite{ref16} with distribution-free risk control, reporting the calibration error on the operational mixed distribution over the four held-out test seeds (n = 440). We ablate the structured components by removing the differential head (leak hypotheses only), by loosening the contract to existence-and-residual only, by disabling the Occam penalty, and by replacing the discriminative active-sensing choice with a random reveal of the same number of hidden nodes; ablation arms share identical scenarios per seed. Finally, we run a fully dual-language-model configuration in which a gpt-oss-120b executor planner proposes the diagnostic plan and an independent deepseek-v4-pro supervisor audits, the two being distinct model families. The planner returns strict-JSON plans through the model runtime's JSON output constraint; all planner and auditor calls are persisted as transcripts, and because leak hypotheses are always tested exhaustively the planner's non-leak choices cannot degrade localization. We compare this pipeline's accept or abstain decisions against the deterministic pipeline on a mixed sample of 16 events. The leak fit (Eq. S4) searches the full severity grid $\{2,5,10,20,35,50\}$ L\,s$^{-1}$, so no severity is structurally unfittable.

A second, fully in silico network verifies that the protocol itself transfers. KY4 is taken from the public Kentucky research database of water distribution system models \cite{ref28} and used unmodified as distributed with the WNTR package. It is partitioned into 25 Leiden zones with 40 sensors (one per $\sim$24 nodes, half the EXA7 density), placed by a degree-based topological observability proxy: the highest-degree junction of every zone, with the remaining sensors filled by global node degree. This is the same rule used for the City H leg, and it replaces the sensitivity-maximizing placement of the EXA7 configuration because neither utility-derived model carries a real gauge layout to reproduce. A per-zone leak-response library is built once, and the identical scenario generator, noise model ($\sigma$ = 0.05 m), acceptance gate and metrics are applied (Table S12).

The battery applies the same pipeline and contract predicates (thresholds per tier in 'The supervisor: goal contract, abstention, and active sensing') and the same five measurements: forced top-1 zone accuracy, coverage, decision precision on acted events, false dispatches on 50 no-leak controls, and the contract versus a scalar existence threshold at matched coverage (Table 1).

The EXA7 test set is mixed: leaks (5--50 L\,s$^{-1}$; the severity analysis extends to 2 L\,s$^{-1}$), demand anomalies, sensor faults and valve mis-states at sensor noise $\sigma$ = 0.05 m (Fig. 4a).

\subsection*{Real-network transfer: L-Town, the City D register leg and standard-protocol legs}

Transfer is first assessed on the public BattLeDIM L-Town benchmark. The 782-node network is partitioned into 25 zones, the 33 provided pressure gauges are the sensor set, and a leak library is built from the nominal network model only. Each of the 33 ground-truth leaks is diagnosed from the provided SCADA series. The competition organizers generated these series from a separate copy of the network model whose roughness and diameters were perturbed, with white measurement noise added: an independently generated benchmark with known ground truth, distinct from our own scenario generator. The observed signature is a diurnal-matched onset step (the per-hour mean over the 24 h after onset minus the 24 h before), and the residual noise scale is 0.15 m.

The City H municipal model (920 junctions, 25 zones, 38 sensors) and the City D model are additionally evaluated under the identical standard in-silico protocol of the EXA7 run ($\sigma$ = 0.05 m, five seeds; Table S12). For City D, sensor placement is optimized by a greedy detection-coverage objective computed only over the pre-simulated leak-response library (2,250 cached responses; no register event, node or observation enters the objective), retaining 23 sensors and raising library actionable coverage (clean |$\Delta$p| $\ge$ 0.45 m) from 29.6\% to 34.7\%. The register evaluation is otherwise unchanged: $\sigma$ = 0.15 m, 50 no-leak controls and a 2--50 L\,s$^{-1}$ severity sweep at 120 seeded nodes. For the survey tier, a direct simulation records each register event's change in every district's net inflow (summed over district-boundary links, reservoir feeds credited to the receiving district, sources external to all districts), verifying the mass-balance identity event by event (Results); the same pass computes the pressure ceiling as the maximum clean |$\Delta$p| over all 541 junctions, under both the pipeline's daily-mean convention and a strict any-time bound (both reported, with identical floor counts). Because that statistic is a single-gauge maximum judged against a single observation, whereas the survey tier averages over the work-order window, a second pass (eval/run\_dagang\_ceiling\_matched.py) recomputes the ceiling on matched terms: the matched filter across all 541 junctions, whose signal-to-noise ratio for a known signature is $\|\Delta\mathbf{p}\|_2/\sigma$, evaluated with the same W = min(window, 14) nights of averaging, which divides $\sigma$ by $\sqrt{\phantom{W}}$W. The survey observation model assumes district inflow metering with nightly uncertainty $\sigma$\_f (sensitivity grid 0.05/0.10/0.20 L\,s$^{-1}$) averaged over the event's real work-order window, capped at 14 nights, with one seeded noise realization per event; a dispatch targets the district with the largest observed rise if it clears u$\cdot$$\sigma$\_f/$\sqrt{\phantom{W}}$W, with u = 3 and the 15-district-corrected u = 3.4 both reported, and 50 no-leak control campaigns (five-night windows) accompany every setting. Like the pressures, the flows are twin-simulated; no real SCADA exists for these events.

Two design alternatives were evaluated and not adopted, with full artifacts committed. A discriminability-optimized placement (greedy nearest-centroid zone-classification objective) raised the library classification proxy from 50.0\% to 64.3\% and the twin-fit to 48.0 $\pm$ 1.8\%, but left decision precision statistically unchanged (83.3 $\pm$ 8.6\% versus 83.1 $\pm$ 11.0\%) while cutting library detection coverage at the 0.45 m floor from 34.7\% to 30.0\%, so the detection-coverage placement is retained. A CUSUM sequential detector for the survey tier, calibrated to the same campaign false-alarm rate, was beaten on district precision at every meter setting with a coverage deficit at the headline setting (40\% at 97\% precision versus the window mean's 44\% at 100\% at $\sigma$\_f = 0.10 L\,s$^{-1}$), consistent with the window mean being the sufficient statistic when the work-order window brackets a persistent leak. Finally, both control sets were analyzed under the same three acceptance presets on the same evidence packages: on the standard leg the transfer gate admits two knife-edge controls (2/50; existence 0.511 and 0.610 against the 0.50 threshold) and the audit preset none (0/50); on the register the transfer and audit outcomes are those reported in the Results (one knife-edge false dispatch and none, respectively; the audit preset is that of the language-model experiments, $\tau$\_e = 0.7, $\delta$ = 0.15, $\alpha$ = 0.20), and adding a $\approx$3$\sigma$ observation floor leaves the one tied control whose observed peak equals the 0.45 m floor.

The register leg uses the City D district network (541 junctions, 475 pipes, 79 throttle-control valves), whose EPANET model was provided by the operating utility and is used unmodified, with its cryptographic checksum recorded alongside the released data. The register is the utility's 2025 repair log of 256 leak work orders. Each order was geocoded from its street address and merged to the nearest numeric junction under an audited mapping: internet-anchored coordinates cross-checked across three candidate coordinate reference systems, per-record anchor confidence and merge distance preserved, and a strict inclusion filter (volume present, connectivity acceptable, non-proxy anchor, node stable across reference systems) that admits 194 orders at 46 junctions. The network is partitioned into 15 zones with 23 sensors (the optimized placement described above); the executor's leak-rate hypothesis grid for this leg extends the standard 2--50 L\,s$^{-1}$ grid down to the register's severity range (0.05/0.1/0.25/0.5/1 L\,s$^{-1}$ added), chosen from the register's rate quantiles alone so the hypothesis space covers the data; forced accuracy is the retrieval localizer's top district (always defined), with a twin-fit forced variant recorded in the released results files. Each event's signature is twin-simulated at its audited node with $\sigma$ = 0.15 m noise; 50 no-leak controls and a 2--50 L\,s$^{-1}$ severity sweep at 120 seeded nodes complete the protocol. Four limits are explicit. The model's topology and equipment are of 2016 vintage; its nodal demands, however, are updated to the 2025 level by a documented nine-year ratio (1.168459, annualized from the city's official city-wide urban tap-water sales 2016--2024, applied uniformly to all 118 demand records with the spatial pattern preserved and equipment screened unchanged), a proxy update that we verified record-by-record against the utility model and that is not a pressure or flow calibration. A demand-ratio sensitivity sweep over the workbook's own band (0.8--1.3 on the 2016 base, with the adopted ratio as regression anchor) leaves the excavation ceiling roughly two orders of magnitude below the actionability floor at every ratio (median 0.0016--0.0025 m; at most 5 of 194 events clear 0.15 m and exactly one clears 0.45 m) and the survey-tier detection count exactly invariant (85 of 194 under the deterministic sizing rule, at every ratio), ruling out demand miscalibration as a cause of the register-leg limits (Table S15). The coordinate datum is unknown, so the mapping is cross-checked across candidate systems rather than asserted. And the work-order window is a reporting proxy, not a measured leak duration.

Throughout, every reported value is written directly to a results file by the run scripts and read back for figures and tables. No metric is hand-entered, hard-coded, or post-edited, and the study inherits no quantities from prior work, an integrity discipline that the auditable architecture is designed to enforce.

The L-Town series are SCADA-format pressure series produced by the benchmark's L-Town data-generating model; each leak's signature is a diurnal-matched onset step.

Each order was geocoded and merged to a model junction under an audited mapping, cross-checked across three coordinate reference systems, with a strict inclusion filter admitting 194 orders at 46 junctions; node assignments therefore reflect audited real locations. Sensor placement, by contrast, is a simulation assumption, optimized from the pre-simulated leak library only, never from register information. The model's 2016 vintage and its remaining mapping limits are set out in this section.

\subsection*{Statistics and reproducibility}

The system is implemented in Python against a WNTR 1.4 \cite{ref41} / EPANET 2.2 \cite{ref42} hydraulic engine. All hydraulic solves are memoized and a per-zone leak-response library is precomputed once. A complete diagnostic episode therefore requires no online optimization beyond cached look-ups and the targeted twin evaluations of Eqs. S4 and S6. On a desktop CPU (Intel, 32 logical cores) the one-time response-library warm-up takes about 4 minutes for EXA7. A complete diagnostic episode then takes a median of 0.029 s (90th percentile 0.07 s, n = 109 cache-warm episodes), of which the active-sensing round contributes a median of 0.007 s. The system therefore operates far inside any operational decision window. A degenerate solve, such as a valve closure that disconnects part of the network, is detected by a non-physical-pressure guard and discarded rather than allowed to corrupt the evidence.

The deterministic executor and supervisor are exactly reproducible by construction. The only stochastic element of the core decision, the additive measurement noise, is governed by a fixed seed registry, so scenario generation is bit-reproducible. The language-model auditor and executor planner, served through the ollama runtime's hosted cloud endpoints, are queried at temperature zero with pinned model tags (deepseek-v4-pro:cloud for the auditor, gpt-oss:120b-cloud for the planner) and a strict-JSON output constraint: the runtime client is local, inference is hosted, and exact repetition therefore depends on the pinned tags remaining available; an earlier auditor tag was retired by the provider during this study, which is why the committed transcripts, not the endpoints, are the reproducibility path. Every call is persisted as a transcript, so the accountability metrics can be replayed from the cached transcripts without live model access, and a parse failure defaults conservatively to rejection.

Because the executor's numbers all originate from the twin and Eq. 3, and the supervisor's hard predicates are deterministic code rather than learned judgement, the entire decision path, from observation to authorized dispatch or abstention, is replayable and verifiable. This is the prerequisite for delegating any automated action on regulated water infrastructure.

All statistical tests are two-sided. Uncertainty is reported as mean $\pm$ standard deviation over five independent random seeds, and small-sample proportions carry exact 95\% Clopper--Pearson confidence intervals. No events were excluded from any analysis; scenario generation is governed by a fixed seed registry; blinding was not applicable because no human raters were involved. Every reported value is written to a committed results file by the run scripts and is read from there by the manuscript and figures. An automated integrity check, included in the released code, enforces that no reported metric is a hard-coded literal, and no number, table or figure is inherited from any prior work on these data. During manuscript preparation the authors used general-purpose large-language-model assistants for language editing and formatting. All scientific content, analyses and numbers originate from the committed pipeline outputs. The authors reviewed and verified the entire text and take full responsibility for it. These writing aids are distinct from the language-model components inside the diagnostic system, which are objects of study rather than authoring tools.

\begin{table}[htbp]
\figcap{Table 1}{\textbf{The identical measurement battery applied to all five networks.} Forced top-1 is the twin-fit zone accuracy when forced to answer every event, except in the City D register column, where it is the retrieval localizer's forced top district (always defined; the twin-fit forced variant is 14.4\% and is recorded in the released results files); coverage and decision precision are at each leg's operating point; no-leak controls are 50 pure-noise scenarios per network under the same acceptance gate; the last row compares the goal contract with a scalar existence threshold admitting the same number of events. Sensor noise is $\sigma$ = 0.05 m for the four in-silico columns, the competition's own noise for L-Town and $\sigma$ = 0.15 m for the register replay; both City D columns use the detection-coverage-optimized 23-sensor placement, and the register column diagnoses audited real leak locations with twin-simulated pressures (Methods). Each in-silico column pools 110 events per seed over n = 5 independent random seeds, 550 events in all, and $\pm$ is the standard deviation across the five per-seed values (DDOF= 1). Forced top-1 is the mean of the five per-seed accuracies, whereas coverage, precision on acted events and the contract-versus-scalar row are computed once on all 550 pooled events and therefore carry no seed spread (per-seed spread in Fig. 4 and Fig. S2 for EXA7 and in Table S12 for KY4, City H and City D; threshold sensitivity in Table S11). Precision on acted events in the two real-tier columns is an exact count with a 95\% Clopper-Pearson interval: L-Town 4/4, 39.8-100.0\%; City D register excavation tier 3/5, 14.7-94.7\%; City D register district-survey tier 85/85, 95.8-100.0\%. The other real-tier cells are counts without intervals. The four in-silico columns share one scenario generator; City D additionally carries its repair register (rightmost column). The N/A entry is a data fact, not a protocol omission: L-Town offers no labeled no-leak window from which real controls could be drawn, as at least one of its 33 labeled leaks is active on every one of the benchmark's 723 days (up to 16 simultaneously), and in a network with continuously growing background leakage a dispatch cannot be scored as false. The register column reports both action tiers, excavation and the flow-balance survey (Methods).}
\begingroup
\setlength{\tabcolsep}{3pt}
\scriptsize
\begin{center}
\begin{tabular}{>{\raggedright\arraybackslash}p{2.00cm}>{\raggedright\arraybackslash}p{2.12cm}>{\raggedright\arraybackslash}p{2.12cm}>{\raggedright\arraybackslash}p{2.12cm}>{\raggedright\arraybackslash}p{2.12cm}>{\raggedright\arraybackslash}p{2.12cm}>{\raggedright\arraybackslash}p{2.12cm}}
\hline
Battery item & EXA7 & KY4 & City H & City D & L-Town & City D register \\
\hline
Data tier & in-silico benchmark & in-silico transfer & in-silico transfer & in-silico transfer & third-party benchmark & audited repair register \\
Nodes/\allowbreak{}zones/\allowbreak{}sensors & 381/15/30 & 959/25/40 & 920/25/38 & 541/15/23 & 782/25/33 & 541/15/23 \\
Events & 550 (5 seeds) & 550 (5 seeds) & 550 (5 seeds) & 550 (5 seeds) & 33 leaks & 194 orders \\
Forced top-1 & 81.7 $\pm$ 5.0\% & 88.7 $\pm$ 3.0\% & 66.7 $\pm$ 5.3\% & 44.7 $\pm$ 4.3\% & 15\% & 12\% \\
Coverage & 40.5\% & 34.4\% & 25.6\% & 20.0\% & 12\% & 2.6\% excavation; 44\% survey \\
Precision on acted & 96.0\% & 96.3\% & 91.5\% & 81.8\% & 100\% (4/\allowbreak{}4; CI 40--100\%) & 60\% excavation (3/\allowbreak{}5); 100\% district (85/\allowbreak{}85) \\
No-leak controls & 0/50 & 0/50 & 0/50 & 2/\allowbreak{}50 (audit 0/\allowbreak{}50) & n/a & 1/\allowbreak{}50 (audit 0/\allowbreak{}50) \\
Contract vs scalar & 96.0\% vs 93.3\% & 96.3\% vs 95.2\% & 91.5\% vs 84.4\% & 81.8\% vs 73.6\% & 100\% vs 75\% & 60\% vs 60\% \\
\hline
\end{tabular}
\end{center}
\endgroup
\end{table}

\section*{Data availability}

The EXA7 benchmark network and all of its derived inputs (zoning, sensor placement and scenario fingerprints) are included in the released code, and the complete results underlying every figure and table are released with them; Source Data are provided with this paper. Per-event diagnostic records for all 33 L-Town leaks and all 194 City D work orders, the numeric values behind Tables S1-S15 and the per-case verdicts of the language-model audit experiments are provided as Supplementary Data 1-3. The KY4 network is a public benchmark from the Kentucky research database of water distribution system models \cite{ref28} and is included unmodified as distributed with the WNTR toolkit. The BattLeDIM L-Town benchmark (network model, SCADA series and leak labels) is publicly available from the 2020 competition. The City H municipal model, the City D utility model and the audited leak-to-node mapping for the 2025 repair register (geocoding cross-checked over three coordinate reference systems, with an independent 39-item validation) are available from the corresponding authors on reasonable request; the raw register is restricted and subject to the operating utility's confidentiality approval. Geographic anchors use OpenStreetMap data (\textcopyright{} OpenStreetMap contributors, ODbL).

\section*{Code availability}

The complete system (executor and supervisor agents, anomaly simulators, and every evaluation and figure script), a pinned software environment and the results records behind every reported value are openly available at \url{https://github.com/mutianwei521/leakagent2} ; the release used for this study will be archived on Zenodo with a DOI upon acceptance. A single command regenerates every number and figure, and the EXA7, KY4 and L-Town legs reproduce from public inputs alone; the City H and City D legs additionally require the network models available on request. The language-model experiments are optional and require an Ollama runtime with the pinned model versions; their complete call transcripts are released, so every audit metric can be recomputed without access to any model. An automated integrity check included in the release verifies that no reported metric is entered as a constant, and a second check verifies the tamper-evident acceptance certificate.

\section*{Acknowledgements}

This research was supported by the open fund from the Key Laboratory of Eco-restoration of Regional Contaminated Environment (Shenyang University), Ministry of Education (No. KF-26-11, to T.M.), the Guangdong Basic and Applied Basic Research Foundation (No. 2026A1515011817, to T.M.), the Liaoning Provincial Education Department Fund Project ``Digital twin-driven global resilience assessment model and dynamic simulation research'' (No. 202464252, to T.M.) and the National Natural Science Foundation of China (No. 62301339, to D.X.). The authors thank the operating utility of City D for the network model and the 2025 repair register, and the City H water utility for the municipal network model.

\section*{Author contributions}

T.M. conceived the study, designed and implemented the executor-supervisor system and its released codebase, generated the anomaly scenario corpora, performed the experiments and the formal analysis, and wrote the original draft. T.M. and D.X. acquired the funding. M.Y. and M.X. supervised the project, administered the collaboration, and secured access to the utility network models and repair registers. Y.W. and D.X. contributed to data curation and to the audited leak-to-node register mapping. W.W. and X.Y. contributed to the hydraulic modeling and validation. M.H. and Q.L. contributed to the methodology and to the environmental interpretation of the results. H.Y. and J.L. contributed to the investigation and visualization. All authors reviewed and edited the manuscript and approved the final version.

\section*{Competing interests}

The authors declare no competing interests.

\clearpage

\section*{Supplementary Information}

\setcounter{equation}{0}
\renewcommand{\theequation}{S\arabic{equation}}
\renewcommand{\theHequation}{SI.\arabic{equation}}

\section*{Supplementary Notes}

\subsection*{Note 1. Governing hydraulics}

Under steady-state operation the network obeys conservation of mass at every node,

\begin{equation}
-\sum_{j\in\mathcal{N}(i)} Q_{ij}(t) = d_i(t) + a_i(t),\quad \forall i\in\mathcal{V}
\end{equation}

and the Hazen--Williams head-loss relationship along every pipe,

\begin{equation}
H_i(t)-H_j(t)= r_{ij}\,|Q_{ij}(t)|^{\,n-1}Q_{ij}(t),\quad \forall (i,j)\in\mathcal{E}
\end{equation}

where $Q_{ij}$ is the pipe flow, $d_i$ the nominal demand, $a_i$ an anomalous additional or perturbing demand, $H_i$ the hydraulic head, $r_{ij}$ the pipe resistance, and $n=1.852$ the head-loss exponent. The digital twin (WNTR/EPANET) solves Eqs. (S1)--(S2); the sensor-space deviation these induce, plus field noise, is main-text Eq. (1).

\subsection*{Note 2. Competing-hypothesis space}

Diagnosis is posed over a structured, mutually exhaustive hypothesis space spanning one dominant anomaly per event,

\begin{equation}
\mathcal{H}=\{h_{\mathrm{leak}}(k)\}_{k=1}^{K}\cup\{h_{\mathrm{dem}}(k)\}_{k=1}^{K}\cup\{h_{\mathrm{sen}}(s)\}_{s\in\mathcal{S}^{*}}\cup\{h_{\mathrm{val}}(a)\}_{a\in\mathcal{A}^{*}}\cup\{h_0\}.
\end{equation}

where $h_{\mathrm{leak}}(k)$ is a leak somewhere in zone $C_k$, $h_{\mathrm{dem}}(k)$ a zone-wide demand anomaly in $C_k$, $h_{\mathrm{sen}}(s)$ a fault on sensor channel $s$, $h_{\mathrm{val}}(a)$ a mis-state of asset $a$, and $h_0$ the null hypothesis that no actionable anomaly exists.

\subsection*{Note 3. Leak fit within a zone}

Because the true leak may sit at any junction in a zone, each leak hypothesis is fitted by the best node--rate pair,

\begin{equation}
\boldsymbol{\mu}_{h_{\mathrm{leak}}(k)} = \arg\min_{v\in C_k,\;q\in\mathcal{Q}}\big\lVert \Delta\mathbf{p}^{\mathrm{obs}}-\boldsymbol{\mu}(v,q)\big\rVert_2^2,
\end{equation}

where the minimization runs over all junctions $v\in C_k$ and the discharge grid $\mathcal{Q}=\{2,5,10,20,35,50\}$ L s$^{-1}$, and $\boldsymbol{\mu}(v,q)$ is the twin-predicted sensor response to a leak of rate $q$ at junction $v$. The residual and its Gaussian log-likelihood are main-text Eq. (2).

\subsection*{Note 4. Posterior summary quantities}

From the posterior (main-text Eq. 3) the supervisor reads three quantities,

\begin{equation}
\Pi_{\mathrm{exist}}=\sum_{k}P\big(h_{\mathrm{leak}}(k)\big),\quad \Delta=P(h_{(1)})-P(h_{(2)}),\quad \mathcal{E}_{\mathrm{bits}}=-\sum_{h}P(h)\log_2 P(h),
\end{equation}

where $\Pi_{\mathrm{exist}}$ is the total leak-existence mass, $\Delta$ the margin of the top hypothesis $h_{(1)}$ over the best alternative $h_{(2)}$ of any family, and $\mathcal{E}_{\mathrm{bits}}$ the posterior entropy in bits.

\subsection*{Note 5. Active-sensing discriminability}

When a rejection reflects an unresolved zone ambiguity, the executor scores every hidden node by the spread of the surviving hypotheses' twin-predicted responses and reveals the highest-scoring nodes (three per round in all reported runs), the first of which is

\begin{equation}
x^{*}=\arg\max_{x\notin\mathcal{S}}\Big(\max_{h}\mu_h(x)-\min_{h}\mu_h(x)\Big),
\end{equation}

where $\mathcal{S}$ is the current sensor set and $\mu_h(x)$ the twin-predicted response of surviving hypothesis $h$ at hidden node $x$; the criterion approximates the measurement of maximal expected information gain.

\subsection*{Note 6. Language-model auditor and planner: prompts and strict-JSON templates}

Both prompts below are reproduced verbatim from the implementation (\texttt{agents/\allowbreak{}llm\_client.py}). This section is generated by importing the constants themselves, so the text quoted here cannot drift from the text the models actually receive.

Both models are queried through the ollama runtime's hosted cloud endpoints under the \texttt{format="json"} constraint at temperature zero, with pinned tags \texttt{deepseek-v4-pro:cloud} for the auditor and \texttt{gpt-oss:120b-cloud} for the planner. Every call is persisted as a transcript, so every reported metric can be replayed without model access. An earlier auditor tag (\texttt{deepseek-v3.1:671b-cloud}) was retired by the provider on 15 July 2026 during the study; every reported audit run uses the tags above, and replay from the committed transcripts does not depend on any endpoint remaining available.

\textbf{Auditor system prompt.}

\begin{lstlisting}
You are an INDEPENDENT safety supervisor for a water-distribution-network leak-diagnosis system. You receive ONLY a structured numeric evidence summary (never the diagnostic agent's reasoning). Decide whether the evidence SUFFICIENTLY and CONSISTENTLY supports dispatching a repair crew to the claimed leak partition.
Residual Mahalanobis-per-dof near 1.0 means a hypothesis REPRODUCES the observation; >>3 means it does NOT fit. Reject (reject=true) if ANY hold:
 1. top_hypothesis.type is not 'leak';
 2. leak_existence_probability < 0.7;
 3. margin_over_next_best < 0.15;
 4. top_hypothesis.fit_residual_mahalanobis > 3 (claimed leak does not fit);
 5. any alternative marked status='falsified' but with residual_mahalanobis < 1.5 (it actually fits well -> inconsistent, the diagnosis is not trustworthy).
Otherwise reject=false. Output STRICT JSON: {"reject": true|false, "reason": "<short>"}.
\end{lstlisting}

\textbf{Auditor input.} The user message carries only the structured numeric evidence summary, serialized as JSON with the fields below. The auditor never sees the executor's reasoning, its retrieved cases or the network model:

\begin{lstlisting}
top_hypothesis: {type, partition, posterior, fit_residual_mahalanobis, candidate_region_nodes}
leak_existence_probability
margin_over_next_best
alternatives (up to 5): [{type, partition, posterior, residual_mahalanobis, status}]
demand_period
\end{lstlisting}

\textbf{Auditor output and fail-safe.} The reply must be strict JSON \texttt{\{"reject": true|false, "reason": "<short>"\}}. A missing or unparsable reply defaults conservatively to \texttt{\{"reject": true, "reason": "auditor\_parse\_failure\_default\_safe"\}}, so a model or endpoint failure withholds the dispatch rather than granting one.

\textbf{Planner system prompt} (dual-language-model configuration only).

\begin{lstlisting}
You are the orchestration head of a water-network leak-diagnosis agent. Given a numeric summary of a pressure-deviation observation, decide which NON-LEAK explanations are worth testing alongside the leak hypotheses: a DEMAND anomaly (a few-to-many sensors depressed broadly across a zone), a SENSOR fault (one or two channels anomalous while the rest read ~0), or a VALVE mis-state (large, structured, network-wide shifts). Always allow the leak hypotheses. Keep the rationale to at most 12 words. Output STRICT JSON: {"include_demand": bool, "include_sensor": bool, "include_valve": bool, "rationale": "<<=12 words>"}.
\end{lstlisting}

\textbf{Planner input.} The user message carries only a numeric summary of the observation, serialized as JSON with the fields:

\begin{lstlisting}
n_sensors
n_anomalous_sensors(|dp|>0.2m)
top_sensor_deviations_m: the six largest sensor deviations by magnitude, in metres
max_abs_dp_m
retrieval_top_zones
\end{lstlisting}

\textbf{Planner output and fail-safe.} The reply must be strict JSON \texttt{\{"include\_demand": bool, "include\_sensor": bool, "include\_valve": bool, "rationale": "<=12 words"\}}. A missing or unparsable reply defaults to enabling all three non-leak families, with \texttt{rationale} set to \texttt{planner\_parse\_failure\_default\_all}. Leak hypotheses are enumerated exhaustively regardless of the planner's reply and are never gated by it, so a planner failure can only add falsification work; it can never remove a candidate zone from consideration.

\subsection*{Note 7. Extended statistics for the EXA7 selective evaluation}

The 96.1 $\pm$ 3.3\% decision precision is a false-dispatch rate of 3.9 $\pm$ 3.3\% on acted events. The executor reaches 81.7 $\pm$ 5.0\% zone accuracy because a leak anywhere in a zone fits that zone's response near-exactly; 81.7 $\pm$ 2.6\% comes from twin-fit re-ranking alone, before active sensing (Methods). The paired McNemar test against the forced localizer is pooled over the five seeds' severity-sweep leak set (2--50 L\,s$^{-1}$, n = 600), and is decisive ($\chi^{2}$ = 171, 254 versus 32 discordant events, P < 0.001; Methods).

\subsection*{Note 8. Why the goal contract dominates a scalar existence score}

\textbf{Why the scalar score saturates.} On the representative seed of Fig. 4a, clear-cut and mis-localized events alike push the summed leak mass toward its ceiling, which is why even its strictest swept threshold still admits 45.5\% of events, and why its best precision is only 96\% (annotated in Fig. 4a).

\textbf{What the predicates add.} The margin and alternative-exclusion predicates veto events whose leading leak hypothesis is closely chased by a rival explanation, and the residual predicate vetoes fits that do not physically reproduce the observation. The operating point itself is not tuned to the test set: it is fixed a priori as the maximum-coverage end of the swept residual gate (Methods). In operational terms, per 100 alarms the forced localizer digs 100 times and is wrong 68 times, whereas the contract digs about 40 times and is wrong once or twice; the 254 of 286 discordant leak events are the McNemar pairs of Supplementary Note 7. The same selective behavior under the identical battery transfers to KY4, City H and City D, compared in Table 1 and analyzed in 'Robustness, generalization and auditable autonomy'.

\subsection*{Note 9. Class asymmetries in differential diagnosis and the fate of mis-typed events}

Typing is weakest for the discrete confounders (33/75 valve, 39/75 sensor), and the operational rate of a confounder dispatched as a leak is 1 in 250 (0.4\%) with the differential head against 50 in 250 (20.0\%) without it. The class asymmetries have physical mechanisms. A sensor fault is never read as a leak because the two hypotheses live in different spaces: the fault form explains a few anomalous channels whose hydraulic neighbors read normal, whereas a leak must depress many sensors coherently, and the twin cannot reproduce the one from the other (Methods). A valve closure is the hardest confounder because it genuinely re-routes flow and depresses downstream pressures, yet its errors still land in other non-leak classes. Across all 250 confounders only ten ended with a leak as the top hypothesis, one demand anomaly over-fitted by the leak family and nine valve closures, and the goal contract vetoed nine of the ten. The 53 leaks that escape the leak type are not scattered at random either: 41 are read as zone-wide demand anomalies, hydraulically the nearest explanation (both are unaccounted withdrawals concentrated in one zone), nine as valve mis-states, three as no anomaly, and none as a sensor fault (Supplementary Table S1). Because a non-leak top hypothesis never dispatches a crew, the price of these confusions is paid in coverage, not in wrong excavations, which is the direction a dispatch system should fail in.

\subsection*{Note 10. Ordering of the trained localization baselines}

The baselines are genuine leak localizers trained on the scenario library and evaluated on the identical test leaks. All the others fall below the cosine k-nearest-neighbor classifier, and their ordering is itself diagnostic. The nearest-neighbor classifier is the strongest learner because the pre-simulated library densely covers each zone's signature shape; the graph network sits near the 6.7\% chance rate at this training scale, a caution for data-hungry localizers in a domain where utilities cannot label thousands of real leaks. The executor's margin of close to ten points over the best trained model comes from a different mechanism than pattern coverage: every candidate must reproduce the observation in the twin, so a noise-distorted signature that fools similarity is still filtered by its residual. The same severity gradient re-appears on the City D network (Fig. 5), where it explains the register leg.

\subsection*{Note 11. Calibrated and conformal comparators, label requirements, and predicate-level ablations}

\textbf{Calibration and comparators.} A split-conformal predictor over the residual-derived confidence controls risk as designed. The expected calibration error is 0.048 over four held-out test seeds (n = 440), and the contract's 97.9\% decision precision carries an exact 95\% CI of 94.0--99.6\%. The logistic-calibrated threshold reaches 99.4\% precision (160/161 acted) at 36.6\% coverage, and the existence-only threshold's strictest setting still acts on 202 events (45.9\% coverage) at 92.6\% precision. The gap to the calibrator is small and informative: the contract approaches a calibrator while needing no labels, and returns per-predicate reasons rather than a scalar. The practical reading concerns labels, not decimals. A logistic threshold must be fitted on held-out correctness labels, which no utility possesses for its real leaks, and the conformal guarantee controls average risk rather than justifying any single dispatch; the contract needs neither ingredient.

\textbf{Predicate-level ablations.} Loosening the contract to existence-and-residual inflates the acted set from 391 to 548 events on the same pooled ablation set, and disabling the Occam penalty multiplies demand-to-leak mis-typing nine-fold (9/100 versus 1/100), which the margin and alternative-exclusion predicates then veto (Supplementary Table S5). Read together, the checks carry measurable and complementary weight: the existence threshold alone reaches 92.6\% at matched coverage, margin and alternative exclusion contribute a further five points, and the Occam penalty protects the differential head upstream of the gate, so no single predicate is decorative.

\textbf{The dual-model configuration.} The fully dual-language-model configuration pairs a gpt-oss-120b planner with a deepseek-v4-pro supervisor. Because leak hypotheses are always tested exhaustively, the planner's wording can influence only which confounders receive extra scrutiny, never whether a leak is missed (Methods).

\subsection*{Note 12. Character of the acted and deferred L-Town events}

The 15\% raw accuracy reflects model mis-calibration and one gauge per $\sim$24 nodes. The four acted L-Town leaks carry an exact 95\% CI of 40--100\% on their 100\% decision precision. Every acted event is a strong abrupt leak (onset |$\Delta$p| 0.40--1.15 m) in the correct zone; the weak and incipient leaks near the noise floor are correctly deferred. The modest acted count is the honest reading of what coverage under severe model error can be.

\subsection*{Note 13. Event-level adjudication of the L-Town acceptance sweep}

At matched four-event coverage on the BattLeDIM L-Town benchmark, the goal contract vetoes one high-confidence but poorly separated event (existence 0.98, margin 0.02) that the scalar existence rule dispatches to the wrong zone, and admits a cleanly separated lower-confidence event (existence 0.52) that is correct. Unlike the calibrated-threshold tie on EXA7, the predicates here strictly improve precision at matched coverage. The honest cost is that one correctly localized event is abstained on, because its margin (0.04) falls below the floor. For context, the first-Pareto winners of the 18-team BattLeDIM competition reached true-positive rates of 56.5\% and 65.2\% on the 2019 leaks \cite{ref3}; the comparison is not like-for-like, and the accountable layer is complementary.

\subsection*{Note 14. Forced localization, no-leak controls and the audit preset on the City D register}

The 194 City D orders reach a maximum loss rate of 3.9 L\,s$^{-1}$, and their median clean signature of 0.002 m is read at the 23 sensors of this stiff, well-pressurized network. Forced to answer, the retrieval localizer places the correct district first in 23 of 194 events (12\%, against a 6.7\% chance rate), and in 41 of 194 when a prediction is scored correct for landing in any adjacent district (21\%, against a $\sim$24\% chance rate): forced guessing barely beats chance, and loses to it under the tolerance. The five excavation dispatches carry an exact 95\% CI of 15--95\% on their 60\% acted precision, and four of the five are at loss rates of 0.75 L\,s$^{-1}$ or more. On the 50 no-leak controls the transfer preset raised one knife-edge false dispatch (2\%; existence 0.51 against the 0.50 threshold on a 3$\sigma$ excursion). The stricter audit preset dismisses that control (0/50) and abstains on all 194 register events: where the evidence cannot carry an excavation, the contract refuses one.

\subsection*{Note 15. Controlled severity sweep and standard-protocol leg on City D}

In the controlled severity sweep on the City D network, injected leaks clear the actionability floor near 35 L\,s$^{-1}$, where coverage reaches 33\% at 78--90\% acted precision (Fig. 5d and Supplementary Table S9). Under the standard in-silico protocol the per-seed mean behind the 81.8\% pooled precision is 83.1 $\pm$ 11.0\% (Supplementary Table S12), so the pipeline functions here; it is the register's leak sizes that remove the signal.

\subsection*{Note 16. Pressure ceiling, mass-balance identity and survey-tier meter assumptions}

With a sensor at all 541 junctions just one of the 194 events would clear the 0.45 m floor. The City D pressure ceiling holds under both the daily-mean and any-time conventions (median best signature 0.0023 m and 0.0030 m; Supplementary Table S10). That a leak raises its district's net inflow by the leak rate itself is an identity the twin confirms on every event (median ratio 1.000; noiseless argmax correct in 194 of 194). The survey tier assumes district inflow metering of nightly uncertainty 0.10 L\,s$^{-1}$ averaged over each order's real work-order window (median five nights); sensitivity over meter assumptions spans 29--56\% coverage at 98--100\% precision (Supplementary Table S10). Excavation-tier coverage stays at 2.6\% while the survey tier runs. On the matched terms the daily-mean convention puts 45 of 194 events at 1$\sigma$ and 17 at 3$\sigma$, and the strict any-time convention 52 and 25; the survey tier still recovers 85.

\subsection*{Note 17. Noise, threshold and cross-network sensitivity of the acceptance gate}

\textbf{Noise sweep.} Across the noise sweep of Fig. 4e the headline metrics are stable across seeds (Fig. S2), and the mechanism of the graceful degradation is the one the contract is built for: as noise more than doubles and doubles again, the gate pays almost entirely in coverage, whose 100\%-precision attainment falls from 29.3\% through 12.9\% to 6.9\% (Fig. 4e), while acted precision moves by a few points. Noise does not push the system into wrong digs; it pushes it back toward abstention.

\textbf{Threshold sensitivity.} The 28 settings are varied one at a time, and the lower end of the 89.9--99.1\% band comes from over-tightening the region bound (R\_max = 25 nodes), which strangles the acted set to 8\% coverage: even a mis-set gate fails toward fewer actions, and every setting stays far above the 31.7\% forced base (Supplementary Table S11).

\textbf{Cross-network transfer.} KY4 is 2.5 times EXA7's size at half its sensor density (Fig. S1e); there the twin-fit reaches 88.7 $\pm$ 3.0\% zone accuracy against 81.3 $\pm$ 4.9\% for retrieval, and, tightened to full precision, KY4 retains 14.2 $\pm$ 16.7\% coverage, the wide spread reflecting two seeds on which no residual gate reaches 100\%.

\subsection*{Note 18. Measured effect of active sensing}

Per-seed changes on the 5--50 L\,s$^{-1}$ headline leak set lie between $-$5.0 and +3.3 points around the unchanged 81.67\% mean. On the extended 2--50 L\,s$^{-1}$ ablation set, which includes the weakest leaks and where accuracy rises from 70.5\% to 74.5\%, a random-reveal control (71.3\%) isolates the discriminability criterion (Supplementary Table S5).

\subsection*{Note 19. Rule-following versus judgement in the independent language-model audit}

\textbf{Uniformity and repeatability of the main stress test.} The catch of 16 of 16 corrupted packages carries an exact 95\% CI of 79--100\%, and the 0/16 false-alarm rate a CI of 0--21\%. The catch was uniform across all three corruption classes (unsupported assertion, fabricated exclusion, inflated evidence; Methods), and the auditor was perfectly stable across eight temperature-zero repeats within one audit session (Methods and Supplementary Table S13).

\textbf{Rule-following versus judgement.} Each of those three corruption classes trips one of the five rules written into the auditor's prompt. The four further classes pass every enumerated rule field by field yet cannot jointly hold: a margin exceeding the leak-existence mass, an alternative outranking the top hypothesis, posteriors summing above one, an empty candidate region. The prompt as written catches essentially none of them (0 to 2 of 64, and those by the parse-failure default), exactly like a literal five-rule checker; the added clause, with nothing enumerated, lets the auditor reject 49 of the 64 by explicit judgement (all 48 of the arithmetic classes, one of the sixteen empty-region packages) with no judged false alarm on the 16 genuine packages, and the planner-family model 22 of 64 with none (Supplementary Table S14). The pragmatic empty-region class still largely passes, which is why the language model may only tighten the hard predicates and never replace them.

\subsection*{Note 20. The four ingredients and what the goal contract establishes under each labelling condition}

Our central claim is deliberately not ``a more accurate localizer''. We take as a premise that a simulation-grounded localizer is weak under field-scale noise, and show that the deployable question is \textit{when to act and how to prove it}. The executor--supervisor agent answers it by uniting four ingredients that are individually known but, in combination, new to this domain: a physics twin used as a falsification instrument, explicit differential diagnosis of non-leak confounders, a code-verifiable goal contract with enforced abstention, and an independent auditor that cannot be talked into agreement. What this establishes depends on whether labeled correctness data exist. On EXA7, where labels can calibrate a confidence threshold, the contract \textit{trails} that calibrated threshold by about 1.5 points at matched coverage (97.9\% versus 99.4\%); it is not superior there. On L-Town, where no labels exist (the realistic utility condition), the contract \textit{strictly improved} decision precision over a scalar threshold at matched coverage (100\% versus 75\%) by vetoing a high-confidence but poorly separated event. On the audited City D repair register it excavated only where the evidence sufficed (five orders, three correct) and recovered 44\% of events through its flow-balance survey tier, where a forced localizer's 194 dispatches would have been right 12\% of the time. This positions accountable leak diagnosis within selective prediction and learning-to-defer \cite{ref13,ref15} while supplying the auditable justification that scalar abstention rules lack; the closest agentic system for water automates simulation but does not gate action \cite{ref23}. The contract is training-free, returns per-predicate reasons and a differential cause, and carries hard safety predicates that a learned threshold does not.

\subsection*{Note 21. Three inconsistencies caught by the provenance discipline and the regeneration of every in-silico number}

The provenance discipline caught and corrected three inconsistencies before release, each in a different layer of the architecture. First, an over-flexible leak hypothesis mimicked zone-wide demand anomalies until the Occam penalty stopped it. Second, a hypothesis-keying inconsistency had inverted the posterior on demand-anomaly cases; it was exposed because every term of the posterior traces to a logged tool output. Third, and found last, an active-sensing round rewrote the margin field of the evidence package without redistributing the hypotheses' posteriors, so that packages were no longer self-consistent and predicate G3 no longer measured what Eq. 4 of the main text says it measures. The open-prompt auditor of Table S14 rejected four genuine packages for exactly this arithmetic inconsistency and was right; on the corrected packages the same auditor raises no judged false alarm. After the correction, every in-silico number in this paper was regenerated through the committed pipeline: the EXA7 operating point moved from 97.6\% precision at 37.5\% coverage to the 96.1\% at 40.5\% reported here, and the KY4 leak-response library, which predated a leak-injection correction and had scaled its responses by that network's diurnal demand pattern, was rebuilt, so every KY4 number comes from the rebuilt library. All three corrections predate the reported runs, and every value in the paper is read from the regenerated records. Auditability must be enforced throughout the architecture, not bolted on, which is what safeguards on human oversight of safety-critical AI demand \cite{ref5}.

\section*{Supplementary Figures}

\begin{center}
\includegraphics[width=\textwidth,height=0.75\textheight,keepaspectratio]{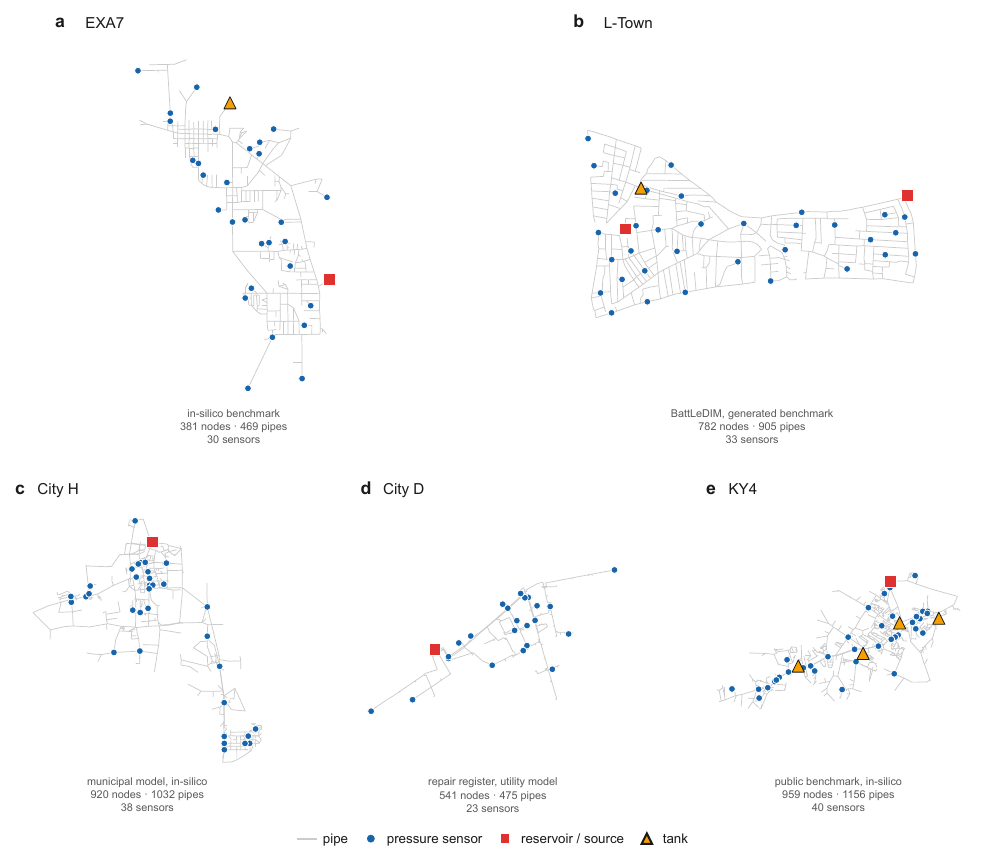}
\end{center}
\figcap{Fig. S1}{\textbf{Topologies of the five validation networks.} Pipe layout drawn from the EPANET INP coordinates for a, EXA7, the in-silico benchmark on which the headline run is performed; b, BattLeDIM L-Town, evaluated against the competition's independently generated SCADA; c, the City H municipal model, a utility-derived network evaluated in silico under the standard protocol; d, City D, the utility's own district-network model behind the register leg; and e, KY4, the public Kentucky-database benchmark, an in-silico network independent of the authors' own models. The panel letters above bind each network to its panel; they do not follow the order in which the networks are introduced in the main text. Pressure-sensor nodes (blue circles), reservoirs and sources (red squares) and tanks (amber triangles) are overlaid on the grey pipe layout; node, pipe and sensor counts are printed beneath each panel. Provenance differs by tier: EXA7, KY4 and City H are evaluated fully in-silico, L-Town uses the competition's independently generated SCADA against the nominal model, and the City D register leg reconstructs each audited repair event in the utility's own model (twin-simulated signatures at the audited real locations; Methods); City D is additionally evaluated fully in-silico under the standard protocol (Table S12), which is why the main text counts four in-silico networks.}

\begin{center}
\includegraphics[width=\textwidth,height=0.75\textheight,keepaspectratio]{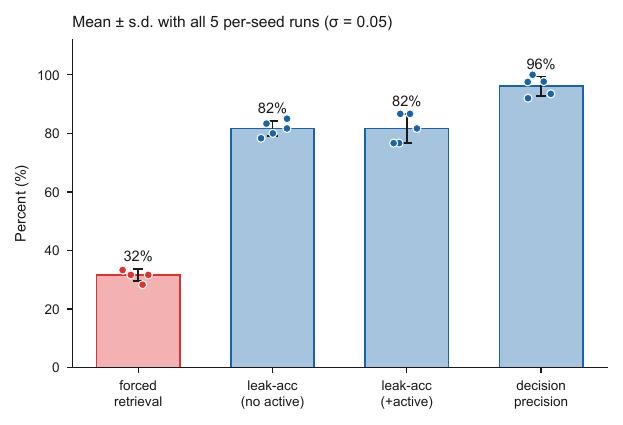}
\end{center}
\figcap{Fig. S2}{\textbf{Robustness of the headline metrics across random seeds (EXA7, $\sigma$ = 0.05 m).} Forced-retrieval top-1 accuracy, leak-partition accuracy without and with supervisor-directed active sensing, and decision precision at the operating point. Bars show the mean over n = 5 independent random seeds, overlaid circles are the individual seeds and error bars are the standard deviation.}

\begin{center}
\includegraphics[width=\textwidth,height=0.75\textheight,keepaspectratio]{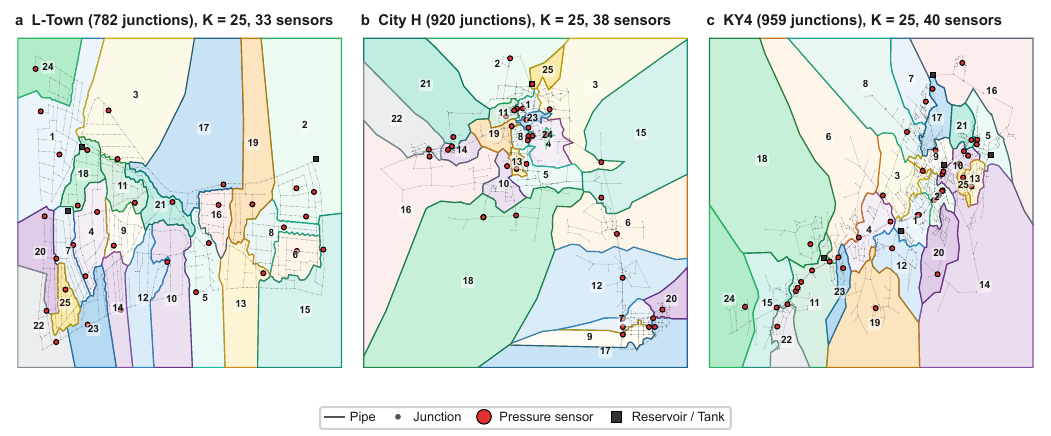}
\end{center}
\figcap{Fig. S3}{\textbf{Hydraulic zoning and sensor placement of the three networks not shown in Fig. 3 (L-Town, City H and KY4).} As Fig. 3 of the main text (Leiden partitions drawn as space-filling territories; colours encode zone identity only): (a) L-Town, K = 25 zones with the 33 pressure gauges of the BattLeDIM benchmark; (b) the City H municipal model, K = 25 zones with 38 sensors; (c) KY4, K = 25 zones with 40 degree-placed sensors. Grey lines are pipes, red dots pressure sensors, dark squares reservoirs and tanks.}

\begin{center}
\includegraphics[width=\textwidth,height=0.75\textheight,keepaspectratio]{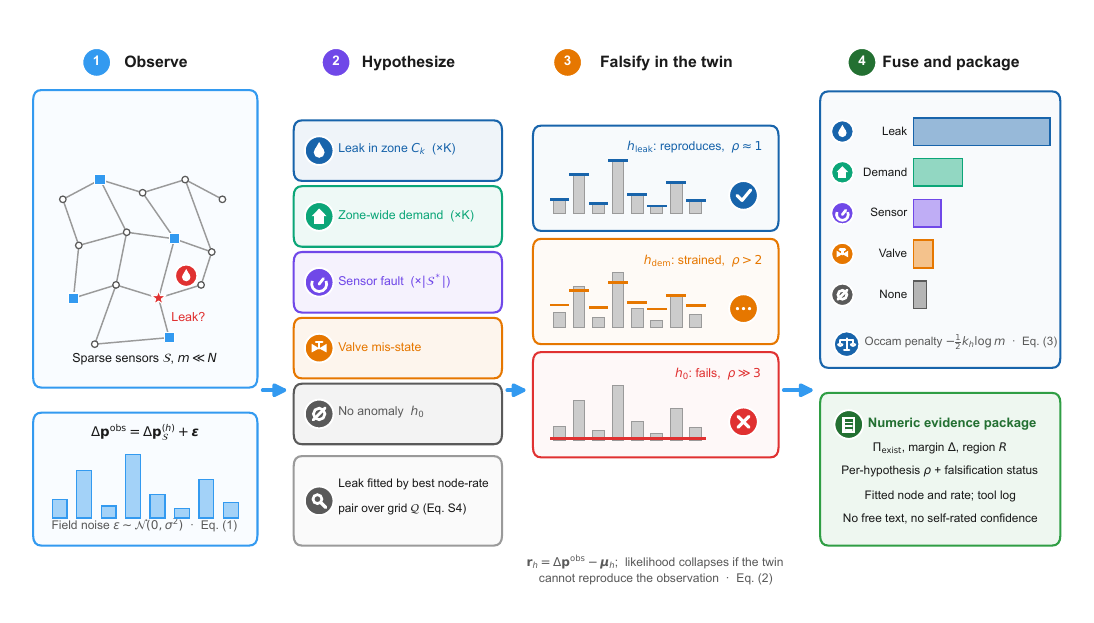}
\end{center}
\figcap{Fig. S4}{\textbf{The executor's diagnostic pipeline (schematic).} A sensor-space observation (Eq. 1; sparse sensors, field noise) is confronted with a mutually exhaustive hypothesis space: a leak in any zone (fitted by its best node-rate pair over the discharge grid, Eq. S4), a zone-wide demand anomaly, a sensor fault, a valve mis-state, or no anomaly. Each hypothesis is realized in the hydraulic digital twin and must reproduce the observation: the residual and its per-degree-of-freedom Mahalanobis distance  (Eq. 2) falsify hypotheses that cannot (red), strain poor fits (amber) and retain reproducing ones (blue). Surviving evidence is fused into a posterior with an Occam penalty against over-flexible families (Eq. 3), from which the leak-existence mass, the top-1 margin and the candidate region are read into a numeric evidence package that carries no free text and no self-rated confidence. Signatures shown are illustrative, not measured values.}

\begin{center}
\includegraphics[width=\textwidth,height=0.75\textheight,keepaspectratio]{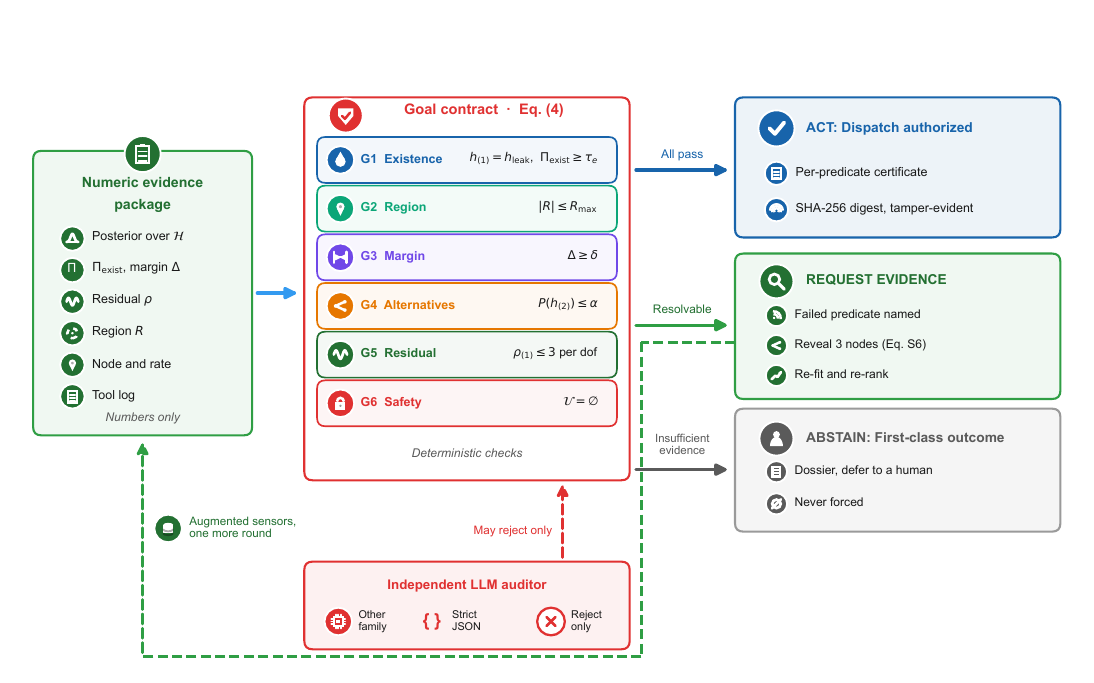}
\end{center}
\figcap{Fig. S5}{\textbf{From evidence to an accountable action (schematic).} The supervisor audits the numeric evidence package (Fig. S4) against the goal contract of Eq. 4: existence, region, margin, alternatives, residual and safety are deterministic arithmetic checks. All checks passing authorizes a dispatch with a certificate that records every predicate's value and pass or fail and is bound to the evidence by a SHA-256 digest; a resolvable failure triggers active sensing, which names the failed predicate, reveals the three most discriminative hidden nodes (Eq. S6) and re-runs the executor on the augmented sensor set (dashed loop); otherwise the system abstains with a dossier and defers to a human. The independent LLM auditor, a different model family at temperature zero, sees only the numeric summary and may add a rejection but can never overturn a hard check.}

\section*{Supplementary Tables}

\figcap{Table S1}{\textbf{Four-class differential-diagnosis confusion matrix on EXA7.} Event counts of the executor's top hypothesis (columns) against the true class (rows), pooled over n = 5 random seeds at $\sigma$ = 0.05 m (600 leak and 250 confounder events); diagonal entries are correct classifications.}

\begingroup
\footnotesize
\begin{center}
\begin{tabular}{lrrrrr}
\hline
True class (rows) /\allowbreak{} Predicted (columns) & leak & demand anomaly & sensor fault & valve misstate & none \\
\hline
\textbf{leak} & 547 & 41 & 0 & 9 & 3 \\
\textbf{demand anomaly} & 1 & 99 & 0 & 0 & 0 \\
\textbf{sensor fault} & 0 & 0 & 39 & 21 & 15 \\
\textbf{valve misstate} & 9 & 6 & 10 & 33 & 17 \\
\textbf{none} & 0 & 0 & 0 & 0 & 0 \\
\hline
\end{tabular}
\end{center}
\endgroup

Sensor faults are never predicted as leaks (leak column, sensor-fault row = 0); demand anomalies are typed 99/100; leaks 547/600.

\figcap{Table S2}{\textbf{Forced top-1 zone accuracy of trained localizers versus the executor's twin-fit on EXA7.} All methods are evaluated on the identical held-out test leaks at $\sigma$ = 0.05 m; values are mean $\pm$ s.d. over n = 5 random seeds.}

\begingroup
\small
\begin{center}
\begin{tabular}{lr}
\hline
Method & Forced Top-1 \\
\hline
GCN (3-layer, trained) & 10.0 $\pm$ 1.2\% \\
MLP (128,64) & 27.7 $\pm$ 1.5\% \\
GraphRAG retrieval & 31.7 $\pm$ 2.0\% \\
RandomForest (200) & 55.7 $\pm$ 5.3\% \\
kNN (k=5, cosine) & 72.0 $\pm$ 3.0\% \\
Executor-Supervisor (twin-fit) & 81.7 $\pm$ 5.0\% \\
\hline
\end{tabular}
\end{center}
\endgroup

The executor--supervisor twin-fit (81.7 $\pm$ 5.0\%, main text) exceeds the strongest trained baseline and, uniquely, abstains and types non-leak causes.

\figcap{Table S3}{\textbf{Leak-partition accuracy as a function of leak severity on EXA7.} Fraction of leaks assigned to the correct topological zone at each injected discharge rate, pooled over n = 5 random seeds at sensor noise $\sigma$ = 0.05 m; n is the number of leak events in each severity band.}

\begingroup
\small
\begin{center}
\begin{tabular}{lrr}
\hline
Leak rate (L\,s$^{-1}$) & n & Leak-partition accuracy \\
\hline
2.0 & 100 & 27\% \\
5.0 & 100 & 60\% \\
10.0 & 100 & 77\% \\
20.0 & 100 & 93\% \\
35.0 & 100 & 93\% \\
50.0 & 100 & 97\% \\
\hline
\end{tabular}
\end{center}
\endgroup

\figcap{Table S4}{\textbf{Split-conformal selective prediction on EXA7.} Realized test risk and coverage at each target risk level, using the residual-derived confidence as the nonconformity score on the mixed test set ($\sigma$ = 0.05 m). The score threshold is fitted on the calibration split, which is the mixed event set of the single development seed 17, and is then applied unchanged to the disjoint test split of n = 440 events pooled over the 4 held-out seeds (42, 101, 2025, 31337); the two splits share no seed and no event. Both columns are exact fractions of those 440 test events, not seed-wise means, so no error bar is defined for them. Distribution-free risk control holds when the realized test risk is at or below the target.}

\begingroup
\small
\begin{center}
\begin{tabular}{lrr}
\hline
Target risk & Realized test risk & Coverage \\
\hline
5\% & 0.6\% & 36.6\% \\
10\% & 5.7\% & 44.1\% \\
\hline
\end{tabular}
\end{center}
\endgroup

Distribution-free risk control holds (realized risk below target at both levels).

\figcap{Table S5}{\textbf{Component ablations on EXA7.} Effect of removing the differential head (a leak-only detector), loosening the goal contract, disabling the Occam/BIC penalty, and replacing the discriminative active-sensing choice with a random reveal, relative to the full system; identical scenarios per seed across arms ($\sigma$ = 0.05 m, pooled over n = 5 random seeds, 850 events per arm, of which 600 are leak events and 250 are confounders, 100 of them demand anomalies). The Effect column mixes denominators, one per quantity, and each cell prints the denominator it uses: confounder false-dispatch counts are out of the 250 confounder events, demand mis-typing out of the 100 demand-anomaly events, and leak-partition accuracy out of the 600 leak events, whereas decision precision is computed only on each arm's own acted cohort, which is 391 events for the full system, 402 without the Occam/BIC penalty, 382 under random reveal and 548 under the loosened contract. Precisions across arms therefore do not share a base, and the acted counts are given with them.}

\begingroup
\setlength{\tabcolsep}{3pt}
\scriptsize
\begin{center}
\begin{tabular}{>{\raggedright\arraybackslash}p{3.27cm}>{\raggedright\arraybackslash}p{12.51cm}}
\hline
Configuration & Effect \\
\hline
Full contract & confounder false-dispatch 1/\allowbreak{}250 \\
No differential head (leak-only) & confounder false-dispatch 50/\allowbreak{}250 \\
Loosened contract (existence + residual only) & decision precision 81.2\% (548 acted, 103 false dispatches) \\
No Occam/\allowbreak{}BIC penalty & demand anomalies mis-typed as leaks 9/\allowbreak{}100 (full system 1/\allowbreak{}100); confounder false-dispatch 1/\allowbreak{}250 (full 1/\allowbreak{}250); decision precision 93.0\% on 402 acted (full 93.6\% on 391 acted) \\
Active sensing: random reveal & leak-partition accuracy 71.3\% (no active sensing 70.5\%, discriminative 74.5\%), each of 600 leak events; decision precision 91.9\% on 382 acted \\
\hline
\end{tabular}
\end{center}
\endgroup

\figcap{Table S6}{\textbf{Per-leak diagnostic outcomes for all 33 BattLeDIM L-Town ground-truth leaks.} Competition-generated SCADA diagnosed against the nominal network model. For each leak: onset type, true zone(s), peak onset signature max|$\Delta$p|, leak-existence posterior, top-one margin, accept/abstain outcome, predicted zone and correctness (a tick marks a correct dispatch).}

\begingroup
\scriptsize
\begin{longtable}{llrrrrlrl}
\hline
Pipe & Type & True zone(s) & max|$\Delta$p| (m) & Existence & Margin & Outcome & Pred. zone & Correct \\
\hline\endfirsthead
\hline Pipe & Type & True zone(s) & max|$\Delta$p| (m) & Existence & Margin & Outcome & Pred. zone & Correct \\\hline\endhead
p257 & incipient & 23 & 0.16 & 0.28 & 0.00 & ABSTAINED & -- &  \\
p461 & incipient & 19 & 0.02 & 0.04 & 0.00 & ABSTAINED & -- &  \\
p232 & incipient & 8,11 & 0.12 & 0.24 & 0.02 & ABSTAINED & -- &  \\
p427 & incipient & 16 & 0.05 & 0.06 & 0.00 & ABSTAINED & -- &  \\
p673 & abrupt & 4 & 1.15 & 1.00 & 1.00 & ACTED & 4 & $\checkmark$ \\
p810 & incipient & 14 & 1.01 & 0.00 & 1.00 & ABSTAINED & -- &  \\
p628 & incipient & 11 & 0.17 & 0.23 & 0.01 & ABSTAINED & -- &  \\
p538 & abrupt & 6,24 & 0.40 & 0.99 & 0.15 & ACTED & 6 & $\checkmark$ \\
p866 & abrupt & 1 & 0.40 & 0.52 & 0.15 & ACTED & 1 & $\checkmark$ \\
p31 & incipient & 0 & 0.16 & 0.01 & 0.00 & ABSTAINED & -- &  \\
p654 & incipient & 11 & 0.23 & 0.02 & 0.04 & ABSTAINED & -- &  \\
p183 & abrupt & 5 & 0.38 & 0.96 & 0.04 & ABSTAINED & -- &  \\
p158 & abrupt & 15 & 0.41 & 0.62 & 0.00 & ABSTAINED & -- &  \\
p369 & abrupt & 2 & 0.43 & 0.74 & 0.05 & ABSTAINED & -- &  \\
p523 & abrupt & 24 & 0.43 & 0.98 & 0.02 & ABSTAINED & -- &  \\
p827 & abrupt & 7 & 0.58 & 1.00 & 0.18 & ACTED & 7 & $\checkmark$ \\
p280 & abrupt & 0 & 0.32 & 0.57 & 0.23 & ABSTAINED & -- &  \\
p653 & incipient & 11 & 0.03 & 0.04 & 0.00 & ABSTAINED & -- &  \\
p710 & abrupt & 15 & 0.15 & 0.28 & 0.01 & ABSTAINED & -- &  \\
p514 & abrupt & 3 & 0.23 & 0.48 & 0.01 & ABSTAINED & -- &  \\
p331 & abrupt & 2 & 0.27 & 0.14 & 0.07 & ABSTAINED & -- &  \\
p193 & incipient & 14 & 0.17 & 0.30 & 0.00 & ABSTAINED & -- &  \\
p277 & incipient & 23 & 0.50 & 0.10 & 0.02 & ABSTAINED & -- &  \\
p142 & abrupt & 11 & 0.40 & 0.35 & 0.46 & ABSTAINED & -- &  \\
p680 & abrupt & 4 & 0.75 & 0.00 & 1.00 & ABSTAINED & -- &  \\
p586 & incipient & 22 & 0.29 & 0.01 & 0.09 & ABSTAINED & -- &  \\
p721 & incipient & 15 & 0.67 & 0.00 & 1.00 & ABSTAINED & -- &  \\
p800 & incipient & 1 & 0.34 & 0.01 & 0.07 & ABSTAINED & -- &  \\
p123 & incipient & 8 & 0.78 & 0.00 & 1.00 & ABSTAINED & -- &  \\
p455 & incipient & 21 & 0.43 & 0.02 & 0.02 & ABSTAINED & -- &  \\
p762 & incipient & 12 & 0.68 & 0.00 & 1.00 & ABSTAINED & -- &  \\
p426 & abrupt & 16 & 0.61 & 0.00 & 1.00 & ABSTAINED & -- &  \\
p879 & incipient & 1,14 & 0.56 & 0.17 & 0.72 & ABSTAINED & -- &  \\
\hline
\end{longtable}
\endgroup

Summary: coverage 12\% (4 acted), decision precision 100\%, forced top-1 zone accuracy 15\%.

\figcap{Table S7}{\textbf{Risk--coverage frontier on BattLeDIM L-Town.} Coverage and decision precision as the leak-existence acceptance threshold is swept over the 33 ground-truth leaks ranked by existence posterior; each row is one distinct operating point on the frontier.}

\begingroup
\small
\begin{longtable}{lrrr}
\hline
Existence threshold & n acted & Coverage & Decision precision \\
\hline\endfirsthead
\hline Existence threshold & n acted & Coverage & Decision precision \\\hline\endhead
1.01 & 0 & 0\% & 100\% \\
1.00 & 2 & 6\% & 100\% \\
0.99 & 3 & 9\% & 100\% \\
0.98 & 4 & 12\% & 75\% \\
0.96 & 5 & 15\% & 80\% \\
0.74 & 6 & 18\% & 67\% \\
0.62 & 7 & 21\% & 57\% \\
0.57 & 8 & 24\% & 50\% \\
0.52 & 9 & 27\% & 56\% \\
0.48 & 10 & 30\% & 50\% \\
0.35 & 11 & 33\% & 45\% \\
0.30 & 12 & 36\% & 42\% \\
0.28 & 13 & 39\% & 38\% \\
0.28 & 14 & 42\% & 36\% \\
0.24 & 15 & 45\% & 33\% \\
0.23 & 16 & 48\% & 31\% \\
0.17 & 17 & 52\% & 29\% \\
0.14 & 18 & 55\% & 28\% \\
0.10 & 19 & 58\% & 26\% \\
0.06 & 20 & 61\% & 25\% \\
0.04 & 21 & 64\% & 24\% \\
0.04 & 22 & 67\% & 23\% \\
0.02 & 23 & 70\% & 22\% \\
0.02 & 24 & 73\% & 21\% \\
0.01 & 26 & 79\% & 19\% \\
0.01 & 27 & 82\% & 19\% \\
0.00 & 33 & 100\% & 15\% \\
\hline
\end{longtable}
\endgroup

At matched coverage (4 acted) the multi-predicate contract reaches 100\% vs 75\% for a scalar existence threshold (it vetoes p523).

\figcap{Table S8}{\textbf{Detectability and outcome by leak-severity band for the City D repair register.} Per-band summary over the 194 audited 2025 work orders (audited real-location node mapping; twin-simulated pressures, $\sigma$ = 0.15 m field noise): number of events, median clean peak signature, forced top-1 zone accuracy, coverage and acted-cohort decision precision. The two right-hand columns run on different denominators. The n column is the number of register events in the band, and forced top-1 and coverage are fractions of that n, but decision precision is a fraction of the acted cohort only, that is, of the events in the band on which the contract dispatched excavation. The acted cohort is given as its own column and is printed inside every precision cell as correct/acted, because it is very small: dispatch occurs in only 3 of the 5 bands, on 1, 1 and 3 events respectively, so those percentages move by whole events and should not be read as rates. Bands with no dispatch have no defined acted precision and are marked n/a. Counts are exact over the register, not means over seeds, so no error bar is defined.}

\begingroup
\footnotesize
\begin{center}
\begin{tabular}{lrrrrrl}
\hline
Leak-rate band (L\,s$^{-1}$) & n & Median clean max|$\Delta$p| (m) & Forced top-1 & Coverage & n acted & Decision precision \\
\hline
{[0,0.05)} & 73 & 0.0003 & 10\% & 0\% & 0 & n/\allowbreak{}a (0 acted) \\
{[0.05,0.1)} & 22 & 0.0008 & 0\% & 5\% & 1 & 0/1 (0\%) \\
{[0.1,0.3)} & 42 & 0.0025 & 10\% & 0\% & 0 & n/\allowbreak{}a (0 acted) \\
{[0.3,1)} & 41 & 0.0095 & 20\% & 2\% & 1 & 1/1 (100\%) \\
>=1 & 16 & 0.0275 & 25\% & 19\% & 3 & 2/3 (67\%) \\
\hline
\end{tabular}
\end{center}
\endgroup

Across the register the median clean signature is 1.9 mm and no event reaches the 0.45 m actionability floor. Under the transfer preset the contract dispatches excavation on 5 of 194 events (60\% acted precision) and raises 2\% false alarms on 50 no-leak controls; the stricter audit preset (existence $\ge$ 0.7) dispatches 0 of 194 and admits 0 of 50 controls. Forced top-1 accuracy is 11.9\%.

\figcap{Table S9}{\textbf{Controlled severity sweep on the City D network.} Single leaks injected at a fixed seeded set of 120 candidate nodes from 2 to 50 L\,s$^{-1}$: median clean peak signature, forced top-1 zone accuracy, coverage and acted-cohort decision precision at each injected rate (twin-simulated pressures with Gaussian field noise of $\sigma$ = 0.15 m, the same field noise level as the register leg of Table S8). Every row is one injected rate over the same 120 nodes, so median signature, forced top-1 and coverage are all on a denominator of 120, but decision precision is a fraction of the acted cohort alone, that is, of the injections at that rate on which the contract dispatched. Those acted cohorts grow with rate and are small at the low end (2 L\,s$^{-1}$: 4, 5 L\,s$^{-1}$: 8, 10 L\,s$^{-1}$: 12, 20 L\,s$^{-1}$: 27, 35 L\,s$^{-1}$: 40, 50 L\,s$^{-1}$: 40), so they are given as their own column and printed inside each precision cell as correct/acted. Counts are exact over the injected set, not means over seeds, so no error bar is defined.}

\begingroup
\footnotesize
\begin{center}
\begin{tabular}{lrrrrr}
\hline
Injected rate (L\,s$^{-1}$) & Median clean max|$\Delta$p| (m) & Forced top-1 & Coverage & n acted & Decision precision \\
\hline
2 & 0.023 & 11\% & 3\% & 4 & 0/4 (0\%) \\
5 & 0.059 & 17\% & 7\% & 8 & 3/8 (38\%) \\
10 & 0.119 & 28\% & 10\% & 12 & 8/12 (67\%) \\
20 & 0.248 & 40\% & 22\% & 27 & 24/27 (89\%) \\
35 & 0.454 & 44\% & 33\% & 40 & 31/40 (78\%) \\
50 & 0.673 & 49\% & 33\% & 40 & 36/40 (90\%) \\
\hline
\end{tabular}
\end{center}
\endgroup

On this stiff, well-pressurized network the median signature clears the noise floor only near 10--20 L\,s$^{-1}$; the entire repair register (Table S8) lies below 4 L\,s$^{-1}$.

\figcap{Table S10}{\textbf{City D register: pressure-information ceiling and the flow-balance survey tier.} Ceiling: with a sensor at every one of the 541 junctions, the median best-possible clean signature over the 194 register events is 0.0023 m; 5 events clear the 0.15 m (1$\sigma$) level and 1 the 0.45 m floor, so a single best-placed gauge reading one daily-mean signature does not make this register pressure-actionable at the assumed noise level (the strict any-time bound gives the same floor counts). This is a bound on the best single sensor, not on every conceivable pressure instrument: an array combining all 541 junctions, or averaging pressure over the same multi-night windows the survey tier uses, clears the floors on more events. Recomputed on those matched terms (the matched filter over all junctions, whose signal-to-noise ratio for a known signature is the l2 norm of the signature over sigma, with the same W = min(window, 14) nights of averaging), 45 of 194 events reach 1 sigma and 17 reach 3 sigma (52 and 25 under the strict any-time convention), still well short of the survey tier's 85. Under a district-adjacency tolerance the forced retrieval prediction lands in or adjacent to the true district in 41/194 events, against a $\sim$24\% chance rate: forced guessing does not beat chance even with the tolerance. Survey tier: district net-inflow deltas are twin-computed per event with source feeds credited to the receiving district (mass-balance identity verified, median ratio 1.000; noiseless argmax correct in 194/194); nightly meter uncertainty $\sigma$\_f is averaged over each order's real work-order window ($\le$14 nights); one seeded noise realization per event; the district with the largest observed rise is survey-dispatched if it clears u$\cdot$$\sigma$\_f/$\sqrt{\phantom{W}}$W. 50 no-leak control campaigns per setting. Symbols and columns of the survey-tier table: $\sigma$\_f is the assumed nightly measurement uncertainty of a district inflow meter, in L\,s$^{-1}$; W is the number of nights averaged for an event, W = min(work-order window in days, 14); and u is the multiple of the resulting standard error $\sigma$\_f/$\sqrt{\phantom{W}}$W that a district's observed inflow rise must clear before a survey is dispatched, reported at u = 3.0 and at the 15-district-corrected u = 3.4. Survey dispatched is the number of the 194 register events dispatched, and coverage is that number as a fraction. District precision is the number of dispatches that named the true district over the number dispatched, with the exact two-sided 95\% Clopper-Pearson interval in parentheses; it is a fraction of the dispatched cohort only, which differs from row to row, and not of the register. Control false alarms is the number of the 50 no-leak control campaigns at that setting on which a survey would have been dispatched. All counts are exact over the register and the control campaigns, so the Clopper-Pearson interval is the only interval reported.}

\begingroup
\setlength{\tabcolsep}{3pt}
\footnotesize
\begin{center}
\begin{tabular}{>{\raggedright\arraybackslash}p{1.53cm}>{\raggedright\arraybackslash}p{1.01cm}>{\raggedright\arraybackslash}p{2.37cm}>{\raggedright\arraybackslash}p{1.11cm}>{\raggedright\arraybackslash}p{5.02cm}>{\raggedright\arraybackslash}p{3.90cm}}
\hline
$\sigma$\_f (L\,s$^{-1}$) & u & Survey dispatched & Coverage & District precision (95\% CI) & Control false alarms (of 50) \\
\hline
0.05 & 3.0 & 109/194 & 56\% & 108/\allowbreak{}109 (99.1\%, 95\% CI 95.0--100.0\%) & 0/50 \\
0.05 & 3.4 & 105/194 & 54\% & 105/\allowbreak{}105 (100.0\%, 95\% CI 96.5--100.0\%) & 0/50 \\
0.10 & 3.0 & 92/194 & 47\% & 91/\allowbreak{}92 (98.9\%, 95\% CI 94.1--100.0\%) & 0/50 \\
0.10 & 3.4 & 85/194 & 44\% & 85/\allowbreak{}85 (100.0\%, 95\% CI 95.8--100.0\%) & 0/50 \\
0.20 & 3.0 & 66/194 & 34\% & 65/\allowbreak{}66 (98.5\%, 95\% CI 91.8--100.0\%) & 0/50 \\
0.20 & 3.4 & 57/194 & 29\% & 57/\allowbreak{}57 (100.0\%, 95\% CI 93.7--100.0\%) & 0/50 \\
\hline
\end{tabular}
\end{center}
\endgroup

Excavation-tier coverage stays at 2.6\% under the transfer preset (zero under the audit preset); the survey tier is the principal recovery path for this register.

\figcap{Table S11}{\textbf{One-at-a-time sensitivity of the goal-contract thresholds on EXA7.} Pooled decision precision and coverage over the extended mixed test set (n = 850 events, five seeds, $\sigma$ = 0.05 m; the same pooled set as Tables S1--S2) as each acceptance threshold is varied around its operating value (existence $\tau$\_e = 0.5, margin $\delta$ = 0.12, alternative-exclusion $\alpha$ = 0.20, residual gate $\rho$\_max = 3, region R\_max = 60 nodes) with the others held at nominal; the per-event evidence rows are generated once and the accept/abstain gate is re-evaluated analytically per setting.}

\begingroup
\small
\begin{longtable}{llrrr}
\hline
Threshold & Value & Decision precision & Coverage & n acted \\
\hline\endfirsthead
\hline Threshold & Value & Decision precision & Coverage & n acted \\\hline\endhead
Existence $\tau$\_e & 0.3 & 93.6\% & 46.0\% & 391 \\
Existence $\tau$\_e & 0.4 & 93.6\% & 46.0\% & 391 \\
Existence $\tau$\_e & 0.5 (nominal) & 93.6\% & 46.0\% & 391 \\
Existence $\tau$\_e & 0.6 & 93.6\% & 46.0\% & 391 \\
Existence $\tau$\_e & 0.7 & 93.6\% & 46.0\% & 391 \\
Existence $\tau$\_e & 0.8 & 93.8\% & 45.9\% & 390 \\
Existence $\tau$\_e & 0.9 & 95.7\% & 44.2\% & 376 \\
Margin $\delta$ & 0.04 & 91.1\% & 47.4\% & 403 \\
Margin $\delta$ & 0.08 & 92.7\% & 46.5\% & 395 \\
Margin $\delta$ & 0.12 (nominal) & 93.6\% & 46.0\% & 391 \\
Margin $\delta$ & 0.16 & 94.6\% & 45.5\% & 387 \\
Margin $\delta$ & 0.2 & 94.8\% & 45.4\% & 386 \\
Margin $\delta$ & 0.24 & 95.3\% & 45.1\% & 383 \\
Alt-exclusion $\alpha$ & 0.1 & 99.1\% & 39.6\% & 337 \\
Alt-exclusion $\alpha$ & 0.15 & 97.7\% & 41.6\% & 354 \\
Alt-exclusion $\alpha$ & 0.2 (nominal) & 93.6\% & 46.0\% & 391 \\
Alt-exclusion $\alpha$ & 0.25 & 92.3\% & 48.9\% & 416 \\
Alt-exclusion $\alpha$ & 0.3 & 91.2\% & 50.7\% & 431 \\
Residual $\rho$\_max & 2.0 & 93.8\% & 45.9\% & 390 \\
Residual $\rho$\_max & 2.5 & 93.6\% & 46.0\% & 391 \\
Residual $\rho$\_max & 3.0 (nominal) & 93.6\% & 46.0\% & 391 \\
Residual $\rho$\_max & 3.5 & 93.6\% & 46.0\% & 391 \\
Residual $\rho$\_max & 4.0 & 93.6\% & 46.0\% & 391 \\
Region R\_max (nodes) & 25 & 89.9\% & 8.1\% & 69 \\
Region R\_max (nodes) & 40 & 93.2\% & 38.0\% & 323 \\
Region R\_max (nodes) & 60 (nominal) & 93.6\% & 46.0\% & 391 \\
Region R\_max (nodes) & 80 & 93.6\% & 46.0\% & 391 \\
Region R\_max (nodes) & 100 & 93.6\% & 46.0\% & 391 \\
\hline
\end{longtable}
\endgroup

Across all 28 one-at-a-time settings, pooled decision precision spans 89.9--99.1\% (nominal 93.6\%).

\figcap{Table S12}{\textbf{Standard in-silico protocol on the three transfer networks (City H, City D and KY4).} Headline metrics under the same scenario generator, noise model ($\sigma$ = 0.05 m) and metrics as the EXA7 headline run; mean $\pm$ s.d. over n = 5 random seeds (110 mixed events per seed). City H: 921 partitioned nodes, 25 zones, 38 sensors. City D: 542 partitioned nodes, 15 zones, 23 sensors placed by the greedy detection-coverage program, whose objective uses only the pre-simulated leak library and raises library actionable coverage (clean |$\Delta$p| $\ge$ 0.45 m) from 29.6\% to 34.7\%. KY4 (public Kentucky benchmark): 959 junctions, 1156 pipes, 25 zones, 40 degree-placed sensors (one per $\sim$24 junctions, half the EXA7 density). The size counts are not defined identically across the three networks and are labeled accordingly: the two utility columns report every node the partitioner assigned to a zone, which includes the network's source node as well as its junctions, whereas the KY4 column reports the model's junction count. The junction counts are the quantity reported in Table 1, the Methods and Fig. 3 of the main text, and are the figure to compare across networks.}

\begingroup
\small
\begin{center}
\begin{tabular}{lrrr}
\hline
Metric & City H & City D & KY4 \\
\hline
Forced retrieval Top-1 & 36.0 $\pm$ 4.2\% & 44.7 $\pm$ 4.1\% & 81.3 $\pm$ 4.9\% \\
Leak-partition accuracy (no active sensing) & 61.0 $\pm$ 5.5\% & 43.3 $\pm$ 3.9\% & 86.7 $\pm$ 5.8\% \\
Leak-partition accuracy (with active sensing) & 66.7 $\pm$ 5.3\% & 44.7 $\pm$ 4.3\% & 88.7 $\pm$ 3.0\% \\
Decision precision at the operating point & 91.5 $\pm$ 3.8\% & 83.1 $\pm$ 11.0\% & 96.4 $\pm$ 2.9\% \\
Coverage at the operating point & 25.6 $\pm$ 1.5\% & 20.0 $\pm$ 3.7\% & 34.4 $\pm$ 1.7\% \\
Coverage at 100\% precision & 5.1 $\pm$ 7.6\% & 3.3 $\pm$ 2.6\% & 14.2 $\pm$ 16.7\% \\
\hline
\end{tabular}
\end{center}
\endgroup

\figcap{Table S13}{\textbf{Independent language-model audit and dual-model pipeline.} Auditor catch rate on corrupted evidence packages, false-alarm rate on genuine packages and temperature-0 decision stability, together with the fully dual-model configuration, from real ollama calls; per-call transcripts are in \texttt{artifacts/\allowbreak{}llm\_transcripts\_*.jsonl}. Every row is an exact count on a small, fully enumerated set, not an estimate over seeds, so no error bar is defined; each row states its own denominator. The 16 genuine (clean) packages are the accepted EXA7 evidence packages of the deterministic executor. Each of them is corrupted exactly once, the three corruption classes being cycled over the 16 packages, so the corrupted set also has 16 packages and the class denominators are unequal: A, unsupported assertion, n = 6; B, fabricated exclusion, n = 5; C, inflated evidence, n = 5. Supplementary Table S14 re-runs the same three classes on all 16 clean packages each and therefore reports them on a larger base; its in-taxonomy counts are not the counts of this table. Decision stability is the temperature-zero repeat-agreement rate, the fraction of repeats returning the modal decision, averaged over 5 of the clean packages audited 8 times each. The dual-model rows are scored on 16 cases.}

\begingroup
\setlength{\tabcolsep}{3pt}
\scriptsize
\begin{center}
\begin{tabular}{>{\raggedright\arraybackslash}p{9.51cm}>{\raggedright\arraybackslash}p{6.27cm}}
\hline
Quantity & Value \\
\hline
Auditor model & deepseek-v4-pro:cloud \\
Catch rate on corrupted packages (n = 16 corrupted packages, one per clean package) & 100\% (16/16) \\
\hspace{0.8em}\hspace{0.8em}$\cdot$ A, unsupported assertion (n = 6) & 100\% (6/6) \\
\hspace{0.8em}\hspace{0.8em}$\cdot$ B, fabricated exclusion (n = 5) & 100\% (5/5) \\
\hspace{0.8em}\hspace{0.8em}$\cdot$ C, inflated evidence (n = 5) & 100\% (5/5) \\
False-alarm rate on genuine packages (n = 16 clean packages) & 0\% (0/16) \\
Decision stability (8 temperature-0 repeats on each of 5 clean packages, 40 audit calls) & 100\% \\
Dual-model pipeline & gpt-oss:120b-cloud planner + deepseek-v4-pro:cloud auditor \\
Accept/\allowbreak{}abstain agreement with deterministic pipeline (n = 16 cases) & 100\% \\
Correct decisions (deterministic /\allowbreak{} dual-model), n = 16 cases & 14/\allowbreak{}16 (88\%) /\allowbreak{} 14/\allowbreak{}16 (88\%) \\
\hline
\end{tabular}
\end{center}
\endgroup

Example gpt-oss-120b planner outputs (varied, discriminative):

- demand=False, sensor=False, valve=False: ``All deviations modest; no pattern for demand, sensor, or valve.''
- demand=True, sensor=False, valve=True: ``Widespread large negative deviations across nearly all sensors''
- demand=False, sensor=False, valve=True: ``Widespread large negative drops suggest valve mis-state, not isolated faults''
- demand=True, sensor=False, valve=True: ``Broad pressure drop across most sensors suggests demand or valve issue.''

\figcap{Table S14}{\textbf{Does the language-model auditor generalize beyond its enumerated rules?} The auditor prompt (Supplementary Methods S6) enumerates five rejection rules, and each of the three corruption classes of Table S13 trips one of them. This follow-up re-uses the same 16 clean ACCEPTED EXA7 packages (identical selection and seeds) and adds four corruption classes that pass every one of the five rules field by field but are jointly impossible: D, a margin larger than the leak-existence probability; E, an alternative listed with a higher posterior than the top hypothesis; F, posteriors summing to well over one; G, an empty candidate region. Three auditors are compared on the same packages: RULES, the five prompt rules implemented literally in code; the language model with the paper's prompt verbatim (as-is); and the language model with the paper's prompt plus one clause asking it to reject any other internal inconsistency, nothing enumerated (open; the added clause reads: ``the summary is internally inconsistent in ANY other way: quantities that cannot jointly hold in a valid probabilistic evidence summary, or a recommendation that could not be acted on.''). The models are the paper's auditor (deepseek-v4-pro:cloud) and, as a second opinion of a different family, the executor planner's model (gpt-oss:120b-cloud); every call is committed as a transcript. Cells give rejected/total; the clean column is the false-alarm count. A rejection annotated fail-safe is not a judgement by the auditor: it is a call whose reply could not be parsed as strict JSON and which therefore fell back to the conservative default of Supplementary Methods S6, \texttt{\{"reject": true, "reason": "auditor\_parse\_failure\_default\_safe"\}}. Fail-safe rejections are included in the printed rejected/total, and the count of them is given in parentheses after the cell, but they are excluded from the judged counts quoted below the table. Stability is the temperature-zero agreement, the fraction of repeats returning the modal decision, over 3 repeats on eight out-of-taxonomy packages, two per class. Every cell is an exact count over a fully enumerated package set, not an estimate over seeds, so no error bar is defined.}

\begingroup
\setlength{\tabcolsep}{3pt}
\scriptsize
\begin{center}
\begin{tabular}{>{\raggedright\arraybackslash}p{2.58cm}>{\raggedright\arraybackslash}p{1.48cm}>{\raggedright\arraybackslash}p{1.25cm}>{\raggedright\arraybackslash}p{1.48cm}>{\raggedright\arraybackslash}p{1.92cm}>{\raggedright\arraybackslash}p{1.18cm}>{\raggedright\arraybackslash}p{1.33cm}>{\raggedright\arraybackslash}p{1.55cm}>{\raggedright\arraybackslash}p{1.55cm}}
\hline
Auditor & Clean (false alarms) & In-taxonomy A+B+C & D margin > existence & E alternative outranks top & F posteriors > 1 & G empty region & Out-of-taxonomy total & Stability \\
\hline
RULES (five prompt rules, code) & 0/16 & 48/48 & 0/16 & 0/16 & 0/16 & 0/16 & 0/64 & 1.000 (deterministic) \\
deepseek-v4-pro:cloud, prompt as-is & 0/16 & 48/48 & 0/16 & 2/\allowbreak{}16 (2 fail-safe) & 0/16 & 0/16 & 2/64 & 1.000 \\
deepseek-v4-pro:cloud, prompt open & 1/\allowbreak{}16 (1 fail-safe) & 48/48 & 16/16 & 16/16 & 16/16 & 7/\allowbreak{}16 (6 fail-safe) & 55/64 & 1.000 \\
gpt-oss:120b-cloud, prompt as-is & 0/16 & 48/48 & 0/16 & 0/16 & 0/16 & 0/16 & 0/64 & 1.000 \\
gpt-oss:120b-cloud, prompt open & 0/16 & 48/48 & 0/16 & 11/16 & 11/16 & 0/16 & 22/64 & 0.958 \\
\hline
\end{tabular}
\end{center}
\endgroup

Read across the rows. With the paper's prompt verbatim the language model behaves as a rule-follower (the prompt ends ``otherwise reject=false'') and rejects essentially nothing outside the enumerated rules, exactly like the code checker. One added clause, without enumerating any new rule, changes that: deepseek-v4-pro:cloud (as-is): out-of-taxonomy 0/64 by judgement, judged false alarms on the genuine packages 0/16; deepseek-v4-pro:cloud (open): out-of-taxonomy 49/64 by judgement, judged false alarms on the genuine packages 0/16; gpt-oss:120b-cloud (as-is): out-of-taxonomy 0/64 by judgement, judged false alarms on the genuine packages 0/16; gpt-oss:120b-cloud (open): out-of-taxonomy 22/64 by judgement, judged false alarms on the genuine packages 0/16. Rejections marked fail-safe are parse failures defaulting to reject and are not counted as judgement. The stronger auditor therefore supplies a generalizing consistency check that a fixed rule list does not, catching every arithmetic impossibility with no judged false alarm, while the pragmatic empty-region class largely passes; it may only tighten the hard predicates and never replace them. In an earlier round of this test the same open auditor rejected four genuine packages for an arithmetic inconsistency that turned out to be a genuine inconsistency in the packages (Note 21); the run reported here uses the corrected packages.

\textbf{Data provenance.} EXA7, KY4 and City H are simulation benchmarks evaluated under one standard in-silico protocol. L-Town uses the public BattLeDIM SCADA, which the competition organizers generated from a perturbed copy of the network model rather than measured in the field, against the nominal model. The City D register leg uses the utility's own network model and its audited 2025 repair register (real work orders, audited real-location node mapping); pressures and district inflows are twin-simulated at the audited locations (Methods). No quantity is inherited from prior work.

\figcap{Table S15}{\textbf{Demand-ratio sensitivity of the City D register tiers.} The utility model's demands are the 2016 base scaled by the evidence-backed nine-year ratio 1.168459 (Methods). Sweeping the overall demand ratio across the update workbook's own hydraulic-sensitivity band re-runs the full ceiling and flow computation of Table S10 at each level (194 events per ratio; the 1.168459 leg reproduces the committed ceiling artifact and serves as the regression anchor). The excavation ceiling stays roughly two orders of magnitude below the 0.15 m floor and the survey-tier detection count is invariant at every ratio, ruling out demand miscalibration as a cause of the register-leg limits. Detection counts use the deterministic sizing rule (true-district inflow rise $\ge$ u$\cdot$$\sigma$\_f/$\sqrt{\phantom{W}}$W, u = 3.4, $\sigma$\_f = 0.10 L\,s$^{-1}$, W = min(window days, 14)).}

\begingroup
\setlength{\tabcolsep}{3pt}
\footnotesize
\begin{center}
\begin{tabular}{>{\raggedright\arraybackslash}p{2.06cm}>{\raggedright\arraybackslash}p{2.86cm}>{\raggedright\arraybackslash}p{2.06cm}>{\raggedright\arraybackslash}p{1.72cm}>{\raggedright\arraybackslash}p{1.02cm}>{\raggedright\arraybackslash}p{1.02cm}>{\raggedright\arraybackslash}p{2.06cm}>{\raggedright\arraybackslash}p{1.72cm}}
\hline
Ratio on 2016 base & Min baseline pressure (m) & Ceiling median (m) & Ceiling p90 (m) & $\ge$0.15 m & $\ge$0.45 m & Mass-balance ratio & Survey detected \\
\hline
0.8 & 32.59 & 0.0016 & 0.0288 & 3/194 & 1/194 & 1.000 & 85/194 \\
1.0 & 31.31 & 0.0020 & 0.0347 & 4/194 & 1/194 & 1.000 & 85/194 \\
1.1 & 30.59 & 0.0022 & 0.0376 & 5/194 & 1/194 & 1.000 & 85/194 \\
1.168459 (adopted) & 30.05 & 0.0023 & 0.0396 & 5/194 & 1/194 & 1.000 & 85/194 \\
1.3 & 28.83 & 0.0025 & 0.0433 & 5/194 & 1/194 & 1.000 & 85/194 \\
\hline
\end{tabular}
\end{center}
\endgroup

No ratio in the plausible band changes the excavation conclusion or the survey-tier detection count; the demand level is not the binding constraint on the register leg.

\end{document}